\pdfoutput=1

\documentclass[11pt]{article}
\usepackage{ACL2023}
\newcommand{\mbart}{{mBART-large-50}\xspace}
\newcommand{\mtf}{{mT5-base}\xspace}
\newcommand{\ibart}{{IndicBART}\xspace}
\newcommand{\mbarttr}{{mBART-large-50-M2O}\xspace}
\newcommand{\ibarttr}{{IndicBART-M2O}\xspace}
\newcommand{\tfb}{{T5-base}\xspace}
\newcommand{\bartbase}{{BART-base}\xspace}
\newcommand{\tfl}{{T5-large}\xspace}
\newcommand{\bartlarge}{{BART-large}\xspace}
\newcommand{\itop}{IE-mTOP\xspace}
\newcommand{\indictop}{IE-multilingualTOP\xspace}
\newcommand{\indicatis}{IE-multiATIS++\xspace}
\newcommand{\hiorig}{hi$_\text{{IE}}$ \xspace}
\newcommand{\hiatis}{hi$_\text{{O}}$ \xspace}
\newcommand{\himtop}{hi$_\text{{O}}$ \xspace}
\newcommand{\NA}{--}

\usepackage{times}
\usepackage{latexsym}
\usepackage{svg}
\usepackage{amsmath}

\usepackage[T1]{fontenc}

\usepackage[utf8]{inputenc}

\usepackage{microtype}

\usepackage{inconsolata}

\usepackage{paralist}

\usepackage{enumitem}
\usepackage{xspace}

\usepackage{booktabs} 

\usepackage{wasysym} 
\usepackage{array}
\usepackage{multirow}
\usepackage{subfig}
\usepackage{tikz}
\usepackage{pgfplots}
\usepackage{float}
\usepackage{arydshln}
\usepackage{centernot}
\usepackage{bbm}
\usepackage{changepage,threeparttable} 
\usepackage[normalem]{ulem}
\usepackage{tabulary}

\newcommand{\skippedDetails}[1]{}

\definecolor{orange}{rgb}{1,0.5,0}
\definecolor{mdgreen}{rgb}{0.05,0.6,0.05}
\definecolor{mdblue}{rgb}{0,0,0.7}
\definecolor{dkblue}{rgb}{0,0,0.5}
\definecolor{dkgray}{rgb}{0.3,0.3,0.3}
\definecolor{slate}{rgb}{0.25,0.25,0.4}
\definecolor{gray}{rgb}{0.5,0.5,0.5}
\definecolor{ltgray}{rgb}{0.7,0.7,0.7}
\definecolor{purple}{rgb}{0.7,0,1.0}
\definecolor{lavender}{rgb}{0.65,0.55,1.0}

\usepackage{amssymb}
\usepackage{pifont}

\usepackage{amsmath}

\newcommand{\quash}[1]{}

\usepackage[para]{footmisc}
\usepackage{arydshln}

\newcommand{\datasetName}{{\sc IE-SemParse}\xspace}

\title{Evaluating Inter-Bilingual Semantic Parsing for Indian Languages}

\author {
    Divyanshu Aggarwal\textsuperscript{\rm 1\thanks{Equal Contribution}}~,
    Vivek Gupta\textsuperscript{\rm {2*}}~,
    Anoop Kunchukuttan\textsuperscript{\rm {3, 4}}
    \\\textsuperscript{\rm 1}American Express, AI Labs;~
     \textsuperscript{\rm 2}University of Utah;~
    \textsuperscript{\rm 3}Microsoft;~ 
    \textsuperscript{\rm 4}AI4Bharat\\
     divyanshu.aggarwal1@aexp.com;~ vgupta@cs.utah.edu ;~ankunchu@microsoft.com
}

\begin{document}

\maketitle
\begin{abstract}
Despite significant progress in Natural Language Generation for Indian languages (IndicNLP), there is a lack of datasets around complex structured tasks such as semantic parsing. One reason for this imminent gap is the complexity of the logical form, which makes English to multilingual translation difficult. The process involves alignment of logical forms, intents and slots with translated unstructured utterance. To address this, we propose an Inter-bilingual Seq2seq Semantic parsing dataset \datasetName for 11 distinct Indian languages. We highlight the proposed task's practicality, and evaluate existing multilingual seq2seq models across several train-test strategies. Our experiment reveals a high correlation across performance of original multilingual semantic parsing datasets (such as mTOP, multilingual TOP and multiATIS++) and our proposed \datasetName suite.  
\end{abstract}


\section{Introduction}
\label{ref:introduction}



\begin{figure}[!h]
\includegraphics[width=7.8cm]{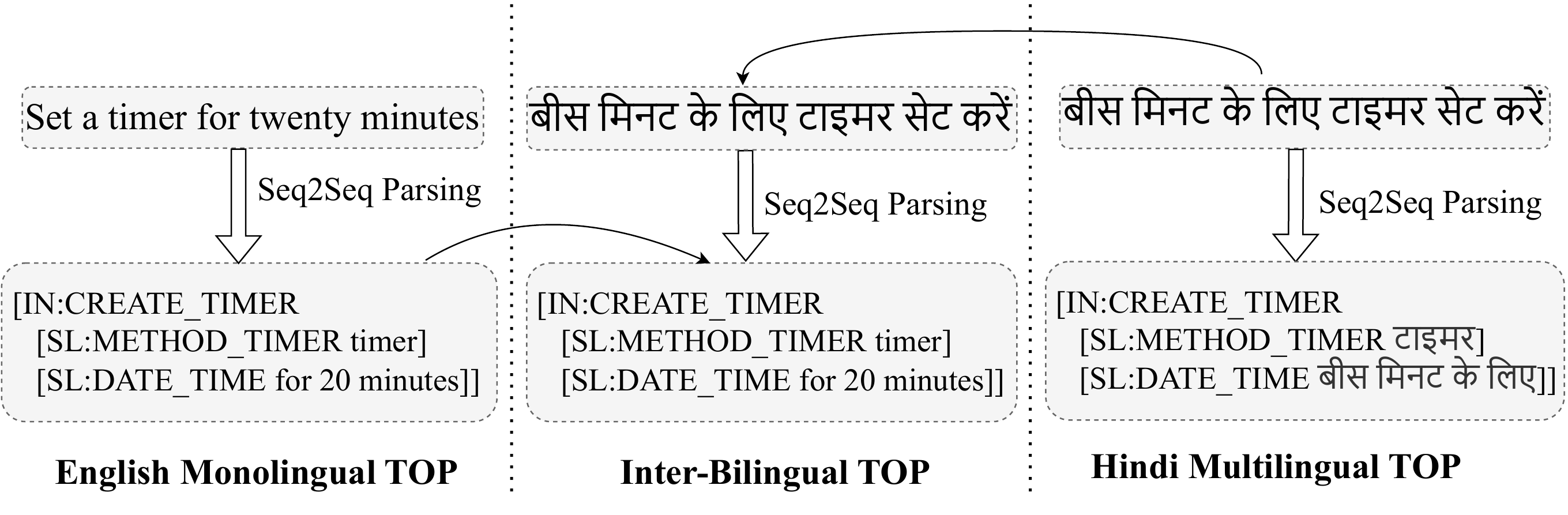}
\vspace{-1.5em}
\caption{\small TOP vs Bilingual TOP.}
\label{fig: top_vs_btop}
\vspace{-1.0em}
\end{figure}

 Task-Oriented Parsing (TOP) is a Sequence to Sequence (seq2seq) Natural Language Understanding (NLU) task in which the input utterance is parsed into its logical sequential form. Refer to Figure \ref{fig: top_vs_btop} where logical form can be represented in form of a tree with intent and slots as the leaf nodes \cite{gupta-etal-2018-semantic-parsing,pasupat-etal-2019-span}. With the development of seq2seq models with self-attention \cite{vaswani2017attention}, there has been an upsurge in research towards developing \emph{generation} models for complex TOP tasks. Such models explore numerous training and testing strategies to further enhance performance \citep{sherborne-lapata-2022-zero, gupta-etal-2022-retronlu}. Most of the prior work focus on the English TOP settings.


 However, the world is largely multilingual, hence new conversational AI systems are also expected to cater to the non-English speakers. In that regard works such as mTOP \citep{li-etal-2021-mtop}, multilingual-TOP \citep{xia-monti-2021-multilingual}, multi-ATIS++ \citep{xu-etal-2020-end,schuster-etal-2019-cross-lingual}, MASSIVE dataset \cite{fitzgerald2022massive} have attempted to extend the semantic parsing datasets to other multilingual languages. However, the construction of such datasets is considerably harder since mere translation does not provide high-quality datasets. The logical forms must be aligned with the syntax and the way sentences are expressed in different languages, which is an intricate process.  




 Three possible scenarios for parsing multilingual utterances exists, as described in Figure \ref{fig: top_vs_btop}. For English monolingual TOP, we parse the English utterance to it's English logical form, where the slot values are in the English language. Seq2Seq models \cite{T5, lewis-etal-2020-bart} tuned on English TOP could be utilized for English specific semantic parsing. Whereas, for multi lingual setting, a \textit{Indic} multilingual TOP (e.g. Hindi Multilingual TOP in Figure \ref{fig: top_vs_btop}) is used to parse Indic utterance to it's respective Indic logical form. Here, the slot values are also Indic (c.f. Figure \ref{fig: top_vs_btop}).\footnote{In both English and Indic Multilingual TOP, the utterance and it's corresponding logic form are in same language, English or Indic respectively.} 


 The English-only models, with their limited input vocabulary, produce erroneous translations as it requires utterance translation. The multilingual models on the other side require larger multilingual vocabulary dictionaries \cite{liang2023xlmv, wang-etal-2019-improving}.  Although models with large vocabulary sizes can be effective, they may not perform equally well in parsing all languages, resulting in overall low-quality output. Moreover, managing multilingual inputs can be challenging and often requires multiple dialogue managers, further adding complexity. Hence, we asked ourselves: \emph{"Can we combine the strengths of both approaches?"}

 Therefore, we explore a third distinct setting: Inter-bilingual TOP. This setting involves parsing Indic utterances and generating corresponding logical forms with English slot values (in comparison, multilingual top has non-english multilingual slot values).  For a model to excel at this task, it must accurately parse and translate simultaneously. The aim of inter-bilingual semantic parsing is to anticipate the translation of non-translated logical forms into translated expressions, which presents a challenging reasoning objective. Moreover, many scenarios, such as e-commerce searches, music recommendations, and finance apps, require the use of English parsing due to the availability of search vocabulary such as product names, song titles, bond names, and company names, which are predominantly available in English. Additionally, APIs for tasks like alarm or reminder setting often require specific information in English for further processing. Therefore, it is essential to explore inter-bilingual task-oriented parsing with English slot values.

 In this spirit, we establish a novel task of Inter-Bilingual task-Oriented Parsing (Bi-lingual TOP) and develop a semantic parsing dataset suite a.k.a \datasetName for Indic languages. The utterances are translated into eleven Indic languages while maintaining the logical structures of their English counterparts.\footnote{Like previous scenarios, the slot tags and intent operators such as METHOD\_TIMER and CREATE\_TIMER are respectively preserved in the corresponding English languages.} We created inter-bilingual semantic parsing dataset \datasetName Suite (IE represents Indic to English). \datasetName suite consists of three Interbilingual semantic datasets namely \itop, \indictop, \indicatis by machine translating English utterances of mTOP, multilingualTOP and multiATIS++ \cite{li-etal-2021-mtop, xia-monti-2021-multilingual, xu-etal-2020-end} to eleven Indian languages described in \S\ref{sec: data creation}. In addition, \S\ref{sec: data creation} includes the meticulously chosen automatic and human evaluation metrics to validate the quality of the machine-translated dataset. 

 We conduct a comprehensive analysis of the performance of numerous multilingual seq2seq models on the proposed task in \S\ref{sec: experiments} with various input combinations and data enhancements. In our experiments, we demonstrate that interbilingual parsing is more complex than English and multilingual parsing, however, modern transformer models with translation fine-tuning are capable of achieving results comparable to the former two. We also show that these results are consistent with those obtained from semantic parsing datasets containing slot values in the same languages as the utterance. Our contributions to this work are the following:


\begin{enumerate}
    \vspace{-5pt}
    \item We proposed a novel task of Inter-Bilingual TOP with multilingual utterance (input) and English logical form (output). We introduced \datasetName, an Inter-Bilingual TOP dataset for 11 Indo-Dravidian languages representing about 22$\%$ of speakers of the world population.
    \vspace{-5pt} 
    \item  We explore various seq2seq models with several train-test strategies for this task. We discuss the implications of an end-to-end model compared to translation followed by parsing. We also compare how pertaining, pre-finetuning and structure of a logical form affect the model performance.
     \vspace{-5pt}
\end{enumerate}

The \datasetName suite along with the scripts will be available at \url{https://iesemparse.github.io/}.



\begin{figure*}[h!]
    {\includegraphics[width=0.95\textwidth]{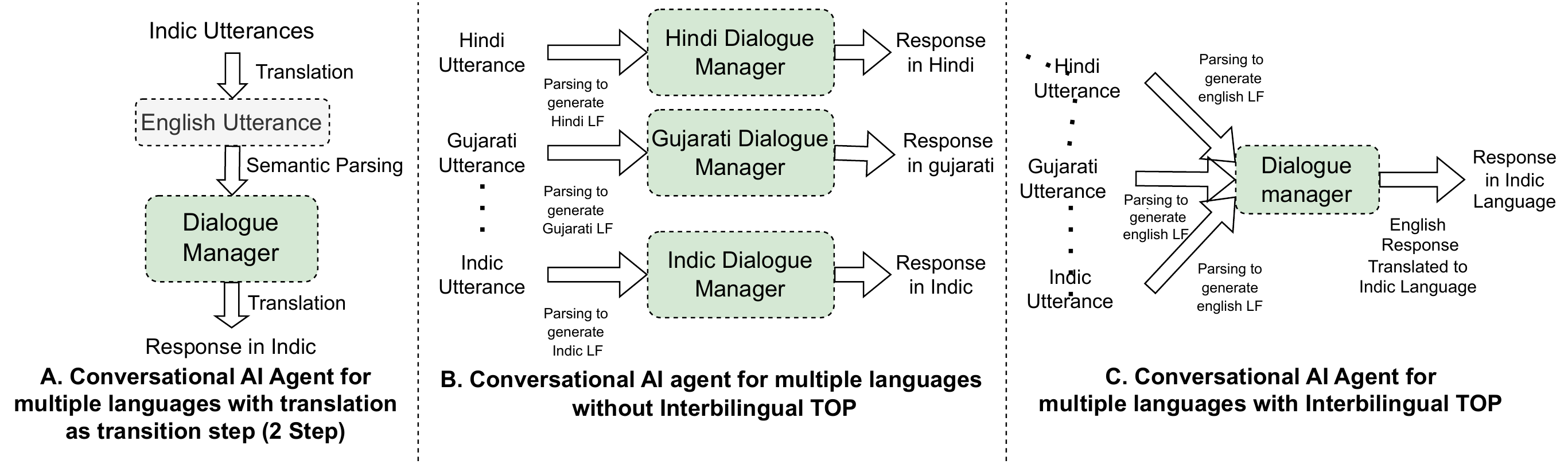}}
    \vspace{-0.75em}
    \caption{\small Conversational AI Agents comparisons with (w/o) inter-bilingual parsing. LF refers to logical form.}
    \vspace{-1.75em}
    \label{fig: motivation}
\end{figure*}

\section{Why Inter Bilingual Parsing?}
\label{sec: task motivation}




In this section, we delve deeper into the advantages of our inter-bilingual parsing approach and how it affects the dialogue management and response generation. We will address the question: \emph{“Why preserve English slot values in the logical form?”}.

\paragraph{Limited Decoder Vocabulary:}


Using only English logical forms simplifies the seq2seq model decoder by reducing its vocabulary to a smaller set. This will make the training process more stable and reduce the chances of hallucination which often occurs in decoders while decoding long sequences with larger vocabulary size \cite{raunak-etal-2021-curious}.


\paragraph{Multi-lingual Models Evaluation:}
In this work, we explore the unique task of translating and parsing spoken utterances into logical forms. We gain valuable insights into the strengths and weaknesses of current multilingual models on this task. Specifically, we investigate how multilingual models compare to monolingual ones, how translation finetuning affects performance, and how the performance of Indic-specific and general multilingual models differ. We also analyze the predictions of the two best models across languages in \S\ref{sec: error analysis}, which is a novel aspect of our task. These insights enhance our understanding of existing multilingual models on \datasetName.



\paragraph{Improved Parsing Latency:}
In figure \ref{fig: motivation}, we illustrate three multilingual semantic parsing scenarios:


\begin{enumerate}
    \vspace{-7pt}
    \item In \textbf{scenario A}, the Indic utterance is translated to English, parsed by an NLU module, and then a dialogue manager delivers an English response, which is translated back to Indic language.
    \vspace{-7pt}
   \item In \textbf{scenario B}, language-specific conversational agents generate a logical form with Indic slot values, which is passed to a language-specific dialogue manager that delivers an Indic response.
    \vspace{-7pt}
    \item In \textbf{scenario C}, a multilingual conversation agent generates a logical form with English slot values, which is passed to an English Dialogue Manager that delivers an English response, which is translated back into Indic language.
    \vspace{-7pt}
\end{enumerate}

\noindent We observe that our approach scenario C is 2x faster than A. We further discuss the latency gains and the performances differences in appendix \S\ref{sec: end 2 end}. Scenario B, on the other hand, has a significant developmental overhead owing to multilingual language, as detailed below.

\paragraph{Handling System Redundancy:} We argue that \datasetName is a useful dataset for developing dialogue managers that can handle multiple languages without redundancy. Unlike existing datasets such as mTOP \cite{li-etal-2021-mtop}, multilingual-TOP \cite{schuster-etal-2019-cross-lingual}, and multi-ATIS++ \cite{xu-etal-2020-end}, which generate logical forms with English intent functions and slot tags but multilingual slot values, our dataset generates logical forms with English slot values as well. This avoids the need to translate the slot values or to create separate dialogue managers for each language, which would introduce inefficiencies and complexities in the system design. Therefore, our approach offers a practical trade-off between optimizing the development process and minimizing the inference latency for multilingual conversational AI agents. Finally, the utilization of a multilingual dialogue manager fails to adequately adhere to the intricate cultural nuances present in various languages \cite{multilingualdm}.

\begin{figure*}[ht!]
    \centering
    \includegraphics[width=0.95\textwidth]{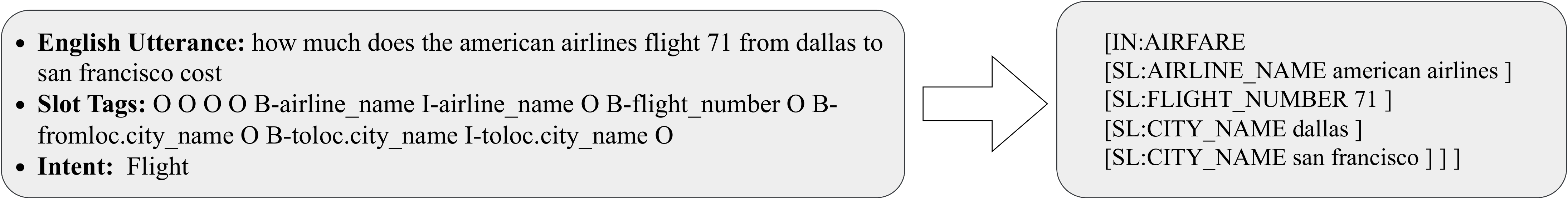}
     \vspace{-0.25em}
    \caption{\small \indicatis Logical Form Generation}
    \label{fig:appendix indicatis lf}
    \vspace{-0.5em}
\end{figure*}

\section{\datasetName Creation and Validation}
\label{sec: data creation}
\vspace{-0.75em}

In this section, we describe the \datasetName creation and validation process in details.
\paragraph{\datasetName Description:}
\begin{table*}[htbp!]\centering
\setlength{\tabcolsep}{5pt}
\small
\begin{tabular}{llrrrrrrrrrrr}
\toprule
\textbf{Score}& \textbf{Dataset} &\textbf{as} &\textbf{bn} &\textbf{gu} &\textbf{hi} &\textbf{kn} &\textbf{ml} &\textbf{mr} &\textbf{or} &\textbf{pa} &\textbf{ta} &\textbf{te} \\
\midrule
&\textbf{Samanantar} &0.83 &0.83 &0.85 &0.87 &0.86 &0.85 &0.85 &0.84 &0.87 &0.87 &0.87 \\
\multirow{2}{*}{\textbf{BertScore}}& \textbf{\itop} &0.83 &0.85 &0.85 &0.87 &0.86 &0.85 &0.86 &0.85 &0.87 &0.87 &0.87 \\
&\textbf{\indictop} &0.98 &0.98 &0.98 &0.96 &0.98 &0.98 &0.99 &0.98 &0.97 &0.98 &0.98 \\
&\textbf{\indicatis} &0.83 &0.85 &0.86 &0.87 &0.86 &0.85 &0.85 &0.85 &0.86 &0.87 &0.87 \\
\hdashline
&\textbf{Samanantar} &0.12 &0.12 &0.11 &0.12 &0.12 &0.12 &0.13 &0.13 &0.12 &0.12 &0.12 \\
\multirow{2}{*}{\textbf{CometScore}}& \textbf{\itop} &0.12 &0.13 &0.12 &0.12 &0.12 &0.13 &0.13 &0.13 &0.14 &0.12 &0.12 \\
&\textbf{\indictop} &0.13 &0.14 &0.14 &0.13 &0.14 &0.14 &0.14 &0.14 &0.14 &0.14 &0.14 \\
&\textbf{\indicatis} &0.13 &0.13 &0.13 &0.13 &0.13 &0.13 &0.13 &0.13 &0.13 &0.13 &0.13 \\
\hdashline
&\textbf{Samanantar} &0.95 &0.96 &0.96 &0.97 &0.96 &0.96 &0.96 &0.96 &0.97 &0.96 &0.96 \\
\multirow{2}{*}{\textbf{BT\_BertScore}}& \textbf{\itop} &0.92 &0.94 &0.93 &0.94 &0.94 &0.93 &0.94 &0.93 &0.93 &0.93 &0.93 \\
& \textbf{\indictop} &0.93 &0.93 &0.89 &0.93 &0.92 &0.96 &0.93 &0.9 &0.92 &0.91 &0.91 \\

&\textbf{\indicatis} &0.91 &0.92 &0.92 &0.93 &0.93 &0.92 &0.92 &0.91 &0.92 &0.92 &0.92 \\

\bottomrule

\end{tabular}
\vspace{-.5em}
\caption{\small Automatic scores on \datasetName and Benchmark Dataset Samanantar.}
\label{tab: automatic scores}
\vspace{-1.5em}
\end{table*}

We create three inter-bilingual TOP datasets for eleven major \emph{Indic} languages that include Assamese (`as'), Gujarat (`gu'), Kannada (`kn'), Malayalam (`ml'), Marathi (`mr'), Odia (`or'), Punjabi (`pa'), Tamil (`ta'), Telugu (`te'), Hindi (`hi'), and Bengali (`bn'). Refer to the appendix \S\ref{sec: appendix further discussions}, for additional information regarding the selection of languages, language coverage of models, and the selection of translation model. The three datasets mentioned  are described below:


\begin{enumerate} 
\vspace{-7pt}
\item \textbf{\itop:} This dataset is a translated version of the multi-domain TOP-v2 dataset. English utterances were translated to Indic languages using IndicTrans \cite{ramesh2021samanantar}, while preserving the logical forms.

\vspace{-7pt}

\item \textbf{\indictop}: This dataset is from the multilingual TOP dataset, where utterances were translated and logical forms were decoupled using the pytext library.\footnote{\url{https://github.com/facebookresearch/pytext}}
\vspace{-7pt}

\item \textbf{\indicatis}: This dataset comes from the multi-ATIS++, where utterances were translated and the logical forms were generated from labelled dictionaries and decoupled, as described in appendix \S \ref{sec: apendix lf atis}.
\vspace{-5pt}
\end{enumerate}

\paragraph{\indicatis Logical Form Creation}
\label{sec: apendix lf atis}
The logical forms are generated from the label dictionaries, where the Intent was labeled with `IN:' tag and Slots were labelled with `SL:' Tags and decoupled like \indictop dataset. 
The process of generating logical forms out of intent and slot tags from the ATIS dataset is illustrated in figure \ref{fig:appendix indicatis lf}.

\vspace{-.5em}

\paragraph{\datasetName Processing:}
To construct \datasetName we perform extensive pre and post processing, as described below:

{\textbf{\emph{Pre-processing}}}
We extensively preprocess \datasetName. We use Spacy NER Tagger\footnote{\url{https://spacy.io/api/entityrecognizer}} to tag date-time and transform them into their corresponding lexical form. E.g. tag date time \emph{``7:30 pm on 14/2/2023."} is transformed to \emph{``seven thirty pm on fourteen february of 2023."}


{\textbf{\emph{Post-processing}}} For many languages some words are commonly spoken and frequently. Therefore, we replace frequently spoken words in \datasetName with their transliterated form, which often sounds more fluent, authentic, and informal than their translated counterparts.



To accomplish this, we replace commonly spoken words with their transliterated form to improve understanding. We created corpus-based transliteration token dictionaries by comparing Hindi mTOP, translated mTOP, and transliterated mTOP datasets. We utilize the human-translated Hindi set of mTOP dataset to filter frequently transliterated phrases and repurpose the same Hindi dictionary to post-process the text for all other Indic languages.


\begin{table*}[!htbp]
\centering
\setlength{\tabcolsep}{5pt}
\small
\begin{tabular}{llrrrrrrrrrrrr}\toprule
\textbf{Dataset} &\textbf{Statistics} &\textbf{as} &\textbf{bn} &\textbf{gu} &\textbf{hi} &\textbf{kn} &\textbf{ml} &\textbf{mr} &\textbf{or} &\textbf{pa} &\textbf{ta} &\textbf{te} \\
\midrule
& \textbf{Human Eval} &3.15 &3.07 &3.65 &4.1 &3.7 &4.12 &4 &4.4 &4.45 &4.03 &3.83 \\
\textbf{\indicatis} & \textbf{Pearson} &0.66 &0.85 &0.69 &0.61 &0.76 &0.62 &0.56 &0.72 &0.61 &0.71 &0.68 \\
& \textbf{Spearman} &0.71 &0.86 &0.42 &0.57 &0.49 &0.51 &0.59 &0.59 &0.59 &0.65 &0.6 \\
\hdashline
& \textbf{Human Eval} &3.06 &3.21 &3.92 &4.46 &4.33 &4.13 &4.24 &4.74 &4.47 &4.22 &3.84 \\
\textbf{\indictop} & \textbf{Pearson} &0.55 &0.79 &0.56 &0.53 &0.45 &0.5 &0.65 &0.42 &0.67 &0.58 &0.59 \\
& \textbf{Spearman} &0.57 &0.74 &0.54 &0.53 &0.45 &0.46 &0.62 &0.63 &0.51 &0.5 &0.49 \\
\hdashline
& \textbf{Human Eval} &3.1 &3.39 &4 &4.42 &4.28 &3.99 &4 &4.61 &4.42 &4.16 &4.13 \\
\textbf{\itop} & \textbf{Pearson} &0.66 &0.74 &0.64 &0.55 &0.61 &0.63 &0.73 &0.45 &0.51 &0.5 &0.62 \\
& \textbf{Spearman} &0.67 &0.7 &0.6 &0.45 &0.4 &0.64 &0.67 &0.41 &0.5 &0.45 &0.5 \\
\bottomrule
\end{tabular}
\vspace{-0.5em}
\caption{\small Human Evaluation Results: \textbf{Human Eval} represents the average score of 3 annotators for each language for each dataset. \textbf{Pearson} is the average pearson correlation of 1st and 2nd, 1st and 3rd and 2nd and 3rd annotators and similarly for \textbf{Spearman} which is spearman correlation.}\label{tab: human eval}
\vspace{-1.75 em}
\end{table*}
\subsection{\datasetName Validation}
As observed in past literature, machine translation can be an effective method to generate high quality datasets \cite{k-etal-2021-analyzing, aggarwal-indicxnli, agarwal-etal-2022-bilingual}. However, due to inherent fallibility of the machine translation system, translations may produce incorrect utterance instances for the specified logical form. Consequently, making the task more complicated and generalizing the model more complex. Thus, it is crucial to examine the evaluation dataset quality and alleviate severe limitations accurately. Early works, including \citet{google-next-thousand, huang-1990-machine, moon-etal-2020-revisiting, quality-estimation-round-trip}, has established that quality estimation is an efficacious method for assessing machine translation systems in the absence of reference data a.k.a the low-resource settings. 

\vspace{-5pt}
\paragraph{Using Quality Estimation:} In our context, where there is a dearth of reference data for the \datasetName translated language, we also determined the translation quality of \datasetName using a (semi) automatic quality estimation technique. Most of recent works on quality estimation compare the results with some reference data and then prove the correlation between reference scores and referenceless quality estimation scores \cite{mitqe, yuan-sharoff-2020-sentence, cuong-xu-2018-assessing}. 
Justifying and interpreting quality estimation metrics, however, remains a stiff challenge for real-world referenceless settings. 

\vspace{-5pt}
\paragraph{\datasetName Automatic Benchmarking:}
 When a parallel corpus in both languages is not available, it is still beneficial to benchmark the data and translation model. In our context, we conducted an evaluation of the Samanantar corpus, which stands as the most comprehensive publicly accessible parallel corpus for Indic languages \citep{ramesh2021samanantar}. The purpose of this assessment was to emulate a scenario wherein the Samanantar corpus serves as the benchmark reference parallel dataset, allowing us to provide a rough estimate of the scores produced by quality estimation models when evaluated in a referenceless setting on a gold standard parallel translation corpus.

 
We use two approaches to compare English and translated text directly. For direct quality estimation of English sentences and translated sentences in a reference-less setting, we utilize Comet Score \cite{rei-etal-2020-comet} and BertScore \cite{bert-score} with XLM-RoBERTa-Large \cite{conneau2020unsupervised} backbone for direct comparison of translated and english utterances. We also calculate BT BertScore \cite{bt_bertscore, moon-etal-2020-revisiting, huang-1990-machine}, which has shown to improve high correlation with human judgement \cite{bt_bertscore} for our three datasets and Samanantar for reference. In this case, we translate the Indic sentence back to English and compare it with the original English sentence using BertScore \cite{bert-score}. The scores for the Samanantar subset on a random subset of filtered 100k phrases and our datasets \datasetName are provided in the table \ref{tab: automatic scores}.

\paragraph{Original vs Machine Translated Hindi:}
\label{sec: appendix human vs machine}
As the human (translated) reference was available in mTOP and multi-ATIS for Hindi language, we leveraged that data to calculate Bert and Comet score to evaluate the translation quality of our machine translation model. We notice a high correlation between both datasets' referenceless and reference scores. Thus suggesting good translation quality for Hindi and other languages.

\begin{table}[!h]\centering
\small
\begin{tabular}{llr}\toprule
\textbf{Dataset} &\textbf{Referenceless Score} &\textbf{Score} \\
\midrule
& \textbf{Comet Score} & 0.83 \\
\textbf{\itop} & \textbf{Bert Score} & 0.96 \\
& \textbf{BT Bert Score} & 0.88 \\

\hdashline

& \textbf{Comet Score} & 0.81 \\
\textbf{\indicatis} & \textbf{Bert Score} & 0.85 \\
& \textbf{BT Bert Score} & 0.87 \\

\bottomrule
\end{tabular}
\vspace{-.5em}
\caption{\small Comet Score, BertScore and BT BertScore of Hindi dataset and translated Hindi dataset for \itop and \indicatis} \label{lab: hindi orig vs trans auto scores}
\vspace{-1.5em}
\end{table}

 In table \ref{lab: hindi orig vs trans auto scores} comet scores and Bert scores are scores keeping original English sentence as source, original Hindi sentence as reference and translated Hindi sentence as hypothesis. For the BT BertScore, the translated Hindi sentence and the original (human-translated) Hindi sentence are back-translated (BT) back onto English and their correlation is assessed using the Bert Score.


\paragraph{\datasetName Human Evaluation:}
In our human evaluation procedure, we employ three annotators for each language \footnote{every annotator was paid 5 INR for each sentence annotation each}. We used determinantal point processes\footnote{\url{https://github.com/guilgautier/DPPy}} \cite{Kulesza_2012} to select a highly diversified subset of English sentences from the test set of each dataset. We select 20 sentences from \indicatis, 120 from \indictop and 60 from \itop. For each dataset, this amounts to more than 1\% of the total test population. We then got them scored between 1-5 from 3 fluent speakers of each Indic English and Indic language by providing them with a sheet with parallel data of English sentences and subsequent translation. 

\textit{Analysis.} We notice that the scores vary with resource variability where languages like ``as'' and ``kn'' have the lowest scores. However, most scores are within the range of 3.5-5 suggesting the high quality of translation for our dataset. Detailed scores are reported in Appendix \S \ref{sec: appendix human eval} table \ref{tab: appendix human eval detailed}.

\section{Experimental Evaluation}
\label{sec: experiments}


\begin{table*}[!h]
\setlength{\tabcolsep}{6pt}
\centering
\footnotesize
\begin{tabular}{m{9em}<{\raggedright}m{10em}<{\raggedright}m{0.4em}m{0.4em}m{0.4em}m{0.4em}m{0.4em}m{0.4em}m{0.4em}m{0.4em}m{0.4em}m{0.4em}m{0.4em}m{0.75em}m{0.75em}m{4.5em}<{\centering}}
\toprule \\[-1.2em]
\multirow{2}{*}{\textbf{Dataset}} & \multirow{2}{*}{\textbf{Model}} & \multicolumn{13}{c}{\textbf{Train All}} & \multirow{2}{*}{\textbf{ModAvg}}
\\ \\[-1.2em]

&&\textbf{as} &\textbf{bn} &\textbf{gu} &\textbf{hi} &\textbf{kn} &\textbf{ml} &\textbf{mr} &\textbf{or} &\textbf{pa} &\textbf{ta} &\textbf{te} &\textbf{hi$_{IE}$} &\textbf{hi$_{O}$}  \\
\midrule \\ \\[-2.3em]
 &\textbf{\ibart} &50 &56 &49 &56 &45 &54 &\textbf{67} &44 &56 &56 &58 & 52 & 60 &50 \\
&\textbf{\mbart} &51 &53 &51 &\textbf{62} &51 &55 &51 &32 &53 &48 &52 & 58 & 66 &51 \\
\textbf{\itop}&\textbf{\mtf} &46 &53 &56 &58 &53 &55 &50 &45 &53 &\textbf{58} &\textbf{58} & 54 & 62 &53 \\
&\textbf{\ibarttr} &54 &57 &57 &\textbf{61} &59 &58 &58 &57 &59 &57 &\textbf{61} & 59 & 63 &58 \\
&\textbf{\mbarttr} &56 &59 &61 &65 &60 &63 &59 &59 &59 &64 &\textbf{65} & 63 & 67 &\textbf{61} \\
\hdashline
& \textbf{Language Average} &51 &56 &55 &\textbf{60} &54 &57 &57 &47 &56 &57 &59 & 57 & 64 &55 \\

\midrule

 & \textbf{\ibart} &44 &50 &57 &\textbf{80} &43 &42 &50 &37 &67 &70 &77 & \NA & \NA &56 \\
& \textbf{\mbart} &44 &57 &66 &\textbf{77} &29 &28 &46 &17 &47 &48 &48 & \NA & \NA  &46 \\
\textbf{\indictop} & \textbf{\mtf} &49 &54 &57 &60 &56 &55 &52 &50 &53 &53 &\textbf{58} & \NA & \NA  &54 \\
&\textbf{\ibarttr} &74 &75 &\textbf{79} &78 &70 &70 &75 &75 &75 &76 &77 & \NA & \NA &\textbf{75} \\
&\textbf{\mbarttr} &54 &57 &60 &\textbf{63} &58 &58 &53 &56 &57 &57 &61 & \NA & \NA &58 \\
\hdashline
& \textbf{Language Average} &51 &56 &55 &\textbf{60} &54 &57 &57 &47 &56 &57 &59 & \NA & \NA &55 \\

\midrule

 & \textbf{\ibart} & 51 &58 &52 &\textbf{70} &50 &41 &63 &25 &50 &39 &56 &66 &76 &54 \\
& \textbf{\mbart} & 54 &\textbf{86} &54 &58 &54 &53 &53 &45 &57 &51 &55 &54 &63 &57 \\
\textbf{\indicatis} & \textbf{\mtf} &67 &\textbf{87} &73 &73 &72 &78 &64 &59 &70 &68 &74 &70 &77 &72 \\
& \textbf{\ibarttr} &70 &\textbf{90} &80 &80 &79 &79 &73 &69 &78 &73 &82 &78 &82 &\textbf{78} \\
& \textbf{\mbarttr}&73 &\textbf{91} &83 &81 &77 &79 &75 &65 &78 &73 &79 &79 &83 &\textbf{78} \\
\hdashline
& \textbf{Language Average} &63 &82 &68 &72 &66 &66 &66 &53 &67 &61 &69 &69 &76 &68 \\

\bottomrule
\end{tabular}
\vspace{-0.35em}
\caption{\small
   $Tree$\_$Labelled$\_$F1*100$ scores for the \textbf{Train All} setting. 
   The bold numbers in the table indicate the row-wise maximum, i.e. the model's best language performance in the given context. The numbers in bold in the \textbf{ModAvg} (Model Average) column indicate the model with the best performance for the train-test strategy specified in the table's heading. Similarly, the numbers in bold in the \textbf{Language Average} row indicate the language with the best performance. Subsequently, hi$_{O}$ refers to the original Hindi dataset from the dataset and hi$_{IE}$ refers to the inter-bilingual dataset constructed by picking Hindi utterances and English logical form and joining them.}

\label{tab: main scores}

\vspace{-0.75em}
\end{table*}
For our experiments, we investigated into the following five train-test strategies: \begin{inparaenum}[1.]
\setlength{\itemsep}{0.1pt}
     \textbf{\item}\textbf{Indic Train:} Models are both finetuned and evaluated on Indic Language.
     \textbf{\item}\textbf{English+Indic Train:} Models are finetuned on English language and then Indic Language and evaluated on Indic language data.
     \textbf{\item}\textbf{Translate Test:} Models are finetuned on English data and evaluated on back-translated English data.
     \textbf{\item}\textbf{Train All:} Models are finetuned on the compound dataset of English + all other 11 Indic languages and evaluated on Indic test dataset.
     \textbf{\item}\textbf{Unified Finetuning:} 
     \ibarttr and \mbarttr models are finetuned on all three datasets for all eleven languages creating unified multi-genre (multi-domain) semantic parsing models for all 3 datasets for all languages. This can be considered as data-unified extension of 4th Setting. 
\end{inparaenum}
 
 \textit{\textbf{Models:}} The models utilized can be categorized into four categories as follows: 
\begin{inparaenum}[(a.)]
\item {\sc Multilingual} such as \textbf{\mbart}, \textbf{\mtf} such as \item {\sc Indic Specific} such as \textbf{\ibart} \item {\sc Translation PreFinetuned} such as \textbf{\ibarttr}, \textbf{\mbarttr}, which are pre finetuned on XX-EN translation task \item {\sc Monolingual (English)} such as \textbf{\tfb}, \textbf{\tfl}, \textbf{\bartlarge}, \textbf{\bartbase} used only in \textbf{Translate Test} Setting. \end{inparaenum} The models are specified in the table's \S\ref{tab: hyper-params} \textit{"Hyper Parameter"} column, with details in the appendix \S\ref{sec: appendix model discussion}. Details of the fine-tuning process with hyperparameters details and the model's vocabulary augmentation are discussed in the appendix \S\ref{sec: appendix hyper parameters} and \S\ref{sec: appendix vocab augment} respectively.


 \textit{Evaluation Metric: } For Evaluation, we use tree labelled F1-Score for assessing the performance of our models from the original TOP paper \citep{gupta-etal-2018-semantic-parsing}. This is preferred over an exact match because the latter can penalize the model's performance when the slot positions are out of order. This is a common issue we observe in our outputs, given that the logical form and utterance are not in the same language. However,  exact match scores are also discussed in appendix \S\ref{sec: appendic exact match}.




 \begin{table*}[htp!]\centering
\small
\begin{tabular}{m{9em}<{\raggedright}m{10em}<{\raggedright}m{0.4em}m{0.4em}m{0.4em}m{0.4em}m{0.4em}m{0.4em}m{0.4em}m{0.4em}m{0.4em}m{0.4em}m{0.4em}m{0.75em}m{0.75em}m{4.5em}<{\centering}}
\toprule \\[-1.2em]
\multirow{2}{*}{\textbf{Dataset}} & \multirow{2}{*}{\textbf{Model}} & \multicolumn{13}{c}{\textbf{Unified Finetuning}} & \multirow{2}{*}{\textbf{ModAvg}}
\\ \\[-1.2em]

& &\textbf{as} &\textbf{bn} &\textbf{gu} &\textbf{hi} &\textbf{kn} &\textbf{ml} &\textbf{mr} &\textbf{or} &\textbf{pa} &\textbf{ta} &\textbf{te}&\textbf{hi$_{IE}$} &\textbf{hi$_{O}$}  \\
\midrule \\ \\[-2.3em]
\multirow{2}{*}{\textbf{\itop}}&\textbf{\ibarttr} &74 &77 &77 &\textbf{81} &79 &78 &78 &77 &79 &77 &81 &79 &83 &78 \\
&\textbf{\mbarttr} &76 &79 &81 &\textbf{85} &80 &83 &79 &79 &79 &84 &85 &83 &87 &\textbf{82} \\
\hdashline
& \textbf{Language Average}  &75 &78 &79 &\textbf{83} &80 &81 &79 &78 &79 &81 &83 &81 &85 &80 \\

\midrule

\multirow{2}{*}{\textbf{\indictop}}&\textbf{\ibarttr} &75 &76 &80 &\textbf{79} &71 &71 &76 &76 &76 &77 &78 &\NA & \NA &\textbf{76} \\
& \textbf{\mbarttr}&55 &58 &61 &\textbf{64} &59 &59 &54 &57 &58 &58 &62 & \NA & \NA &59  \\
\hdashline
& \textbf{Language Average}  &65 &67 &71 &\textbf{72} &65 &65 &65 &67 &67 &68 &70 & \NA & \NA &67 \\ 

\midrule

\multirow{2}{*}{\textbf{\indicatis}}&\textbf{\ibarttr} &80 &80 &90 &\textbf{90} &89 &89 &83 &79 &88 &83 &92 &88 &92 &\textbf{84} \\
&\textbf{\mbarttr} &83 &82 &93 &\textbf{91} &87 &89 &85 &75 &88 &83 &89 &89 &93 &\textbf{84} \\
\hdashline
& \textbf{Language Average} &82 & 82 &92 &\textbf{91} &88 &89 &84 &77 &88 &83 &\textbf{91} &89 &93 & \textbf{84} \\
\bottomrule
\end{tabular}

\vspace{-0.35em}
\caption{\small $Tree$\_$Labelled$\_$F1*100$ scores of \textbf{\ibarttr} and \textbf{\mbart} model trained on all languages and all datasets. Other notations similar to that of Table \ref{tab: main scores}.}

\label{tab: all lang data results}
\vspace{-1.5em}
\end{table*}

\subsection{Analysis across Languages, Models and Datasets}

\label{sec: result analysis}
We report the results of \textbf{Train All} and \textbf{ Unified Finetuning} settings for all datasets in table \ref{tab: main scores} and \ref{tab: all lang data results} in the main paper as these were the best technique out of all. The scores for other train-test strategies such as translate test, Indic Train, English+Indic Train for all 3 datasets are reported in appendix \S\ref{sec: appendix indictop and indicatis} table \ref{tab:itop scores}, \ref{tab: indictop scores} and \ref{tab: english indic scores} respectively. However, we have discussed the comparison between train-test settings in the subsequent paragraphs.

 {\textbf{Across Languages:}} Models perform better on high-resource than medium and low-resourced languages for \textbf{Train All} setting. This shows that the proposed inter-bilingual seq2seq task is challenging. In addition to linguistic similarities, the model performance also relies on factors like grammar and morphology \cite{pires-etal-2019-multilingual}. For other settings such as \textbf{Translate Test}, \textbf{Indic Train}, and \textbf{English+Indic}, similar observations were observed.

{\textbf{Across Train-Test Strategies:}} 
Translate Test method works well, however end-to-end English+Indic and Train All models perform best; due to the data augmentation setting, which increases the training size.\footnote{By 2x (English + Indic) and 12x (1 English + 11 Indic).} However, the benefits of train data enrichment are much greater in {\bf Train All} scenario because of the larger volume and increased linguistic variation of the training dataset.
We also discuss the comparisons in inference latency for a 2-step vs end-to-end model in \S\ref{sec: task motivation}.




 {\textbf{Across Datasets:}} We observe that \indictop is the simplest dataset for models, followed by \itop and \indicatis. This may be because of the training dataset size, since \indictop is the largest of the three, followed by \itop and \indicatis. In addition, \indictop is derived from TOP(v1) dataset which have utterances with more simpler logical form structure (tree depth=1). \itop, on the other hand, is based on mTOP, which is a translation of TOP(v2), with more complex logical form having (tree depth>=2). We discuss the performance of models across logical form complexity in \S\ref{sec: error analysis}. For \textbf{Unified Finetuning} we observe an average performance gain of 0.2 in the tree labelled F1 score for all languages for all datasets as reported in table \ref{tab: all lang data results} in appendix.


{\textbf{Across Models:}} We analyse the performance across various models based on three criteria, language coverage, model size and translation finetuning, as discussed in detail below:

(a.) \textit{\textbf{Language Coverage:}} Due to its larger size, \mbarttr performs exceptionally well on high-resource languages, whereas \ibarttr performs uniformly across all the languages due to its indic specificity. In addition, translation-optimized models perform better than those that are not. \mbart outperforms \mtf despite its higher language coverage, while \mbart's superior performance can be ascribed to its denoising pre-training objective, which enhances the model's ability to generalize for the \emph{"intent"} and \emph{"slot"} detection task. In section \S\ref{sec: error analysis} we discuss more about the complexity of the logical forms.




(b.) \textit{\textbf{Model Size:}} 
While model size has a significant impact on the Translate Test setting for monolingual models, we find that pre-training language coverage and Translation fine-tuning are still the most critical factors.
For example, despite being a smaller model, \ibart outperforms \mtf on average for similar reasons. Another reason for better performance for \ibart and \mbart denoising based seq2seq pre-training vs multilingual multitask objective of \mtf.


(c.) \textit{\textbf{Translation Finetuning:}} 
The proposed task is a mixture of semantic parsing and translation. We also observe this empirically, when models finetuned for translation tasks perform better. This result can be attributed to fact that machine translation is the most effective strategy for aligning phrase embeddings by multilingual seq2seq models \cite{voita-etal-2019-bottom}, as emphasized by \citet{li-etal-2021-mtop}. In addition, we observe that the models perform best in the \textbf{Train All} setting, indicating that data augmentation followed by fine-tuning enhances performance throughout all languages on translation fine-tuned models.

 \textbf{Original vs Translated Hindi:}
 We also evaluated the performance of Hindi language models on original datasets (\himtop) and (\hiorig) which combine Hindi utterances with logical forms of English of mTOP and multi-ATIS++ datasets, as shown in table \ref{tab: main scores}. Inter-bilingual tasks pose a challenge and result in lower performance, but translation-finetuned models significantly reduce this gap. Model performance is similar for both `hi' and `\hiorig', indicating the quality of translations. Additional details can be refered in Appendix \S\ref{sec: appendix original hindi vs translated hindi results}.

\begin{figure}[!h]
\vspace{-0.25em}
    {\includegraphics[width=7.8cm]{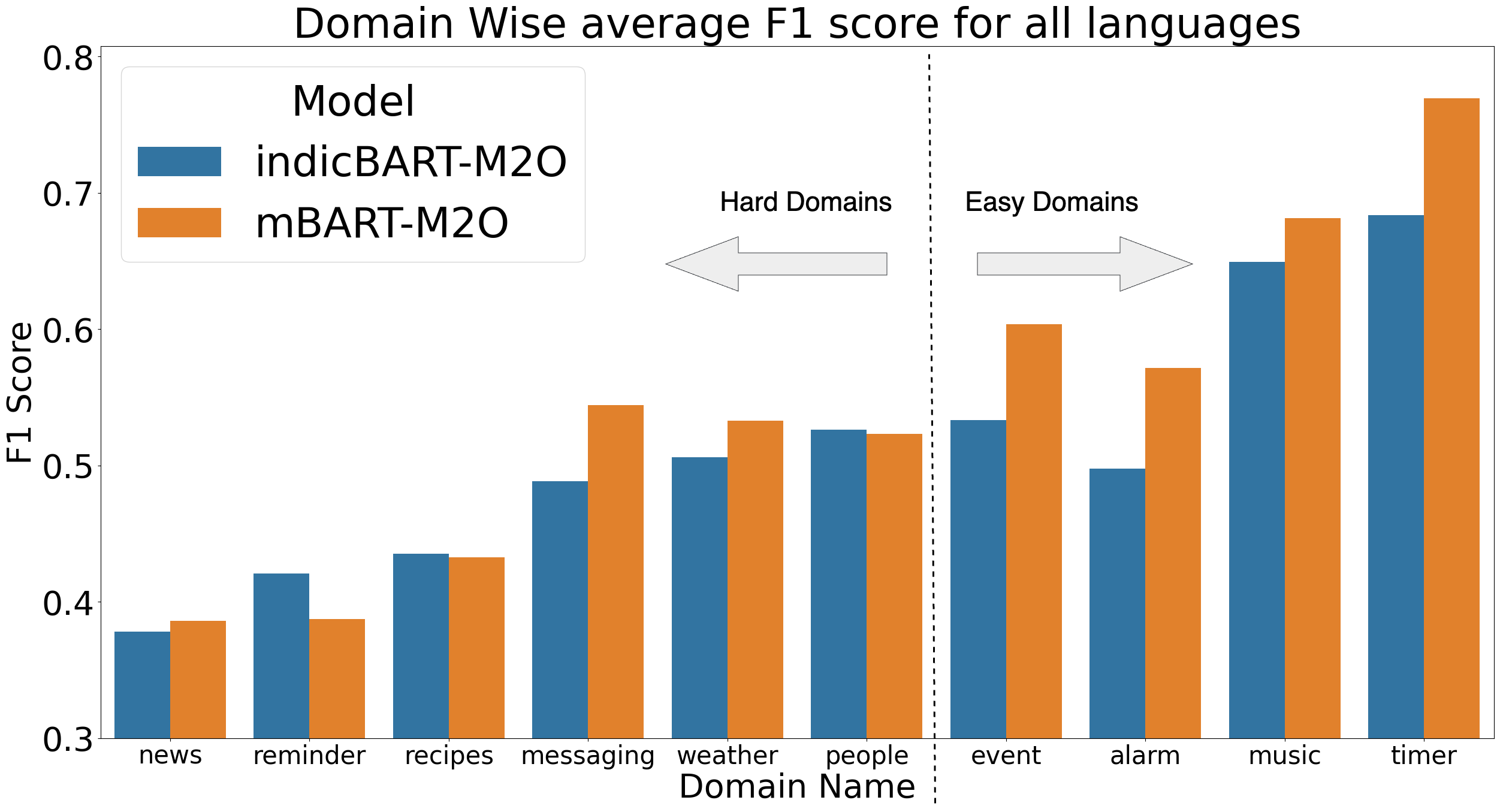}}
    \caption{\small Domain Wise all language average F1 score in \itop dataset for \ibarttr and \mbarttr. 
    }
   \label{fig: domain f1}
    \vspace{-1.50em}
\end{figure}

 \textbf{Domain Wise Comparison:} 
 \itop dataset contains domain classes derived from mTOP. We compare the average F1 scores for different domains in \itop dataset for \ibarttr and \mbarttr in the \textbf{Train All} setting, as shown in Figure \ref{fig: domain f1}. We observe that \mbarttr outperforms \ibarttr for most domains except for people and recipes, where both perform similarly well due to cultural variations in utterances.

\subsection{Analysis on Logical Forms}
\label{sec: error analysis}

In this paper, we maintain the slot values in the English language and ensure consistency in the logical form across languages for each example in every dataset. This can be useful in assessing the model performance across language and datasets on the basis of logical form structure which we have analysed in this section. Previous works have shown a correlation between model performance and logical form structures \citep{gupta-etal-2022-retronlu}.

\begin{figure}[!h]
\vspace{-0.25em}
    {\includegraphics[width=7.8cm]{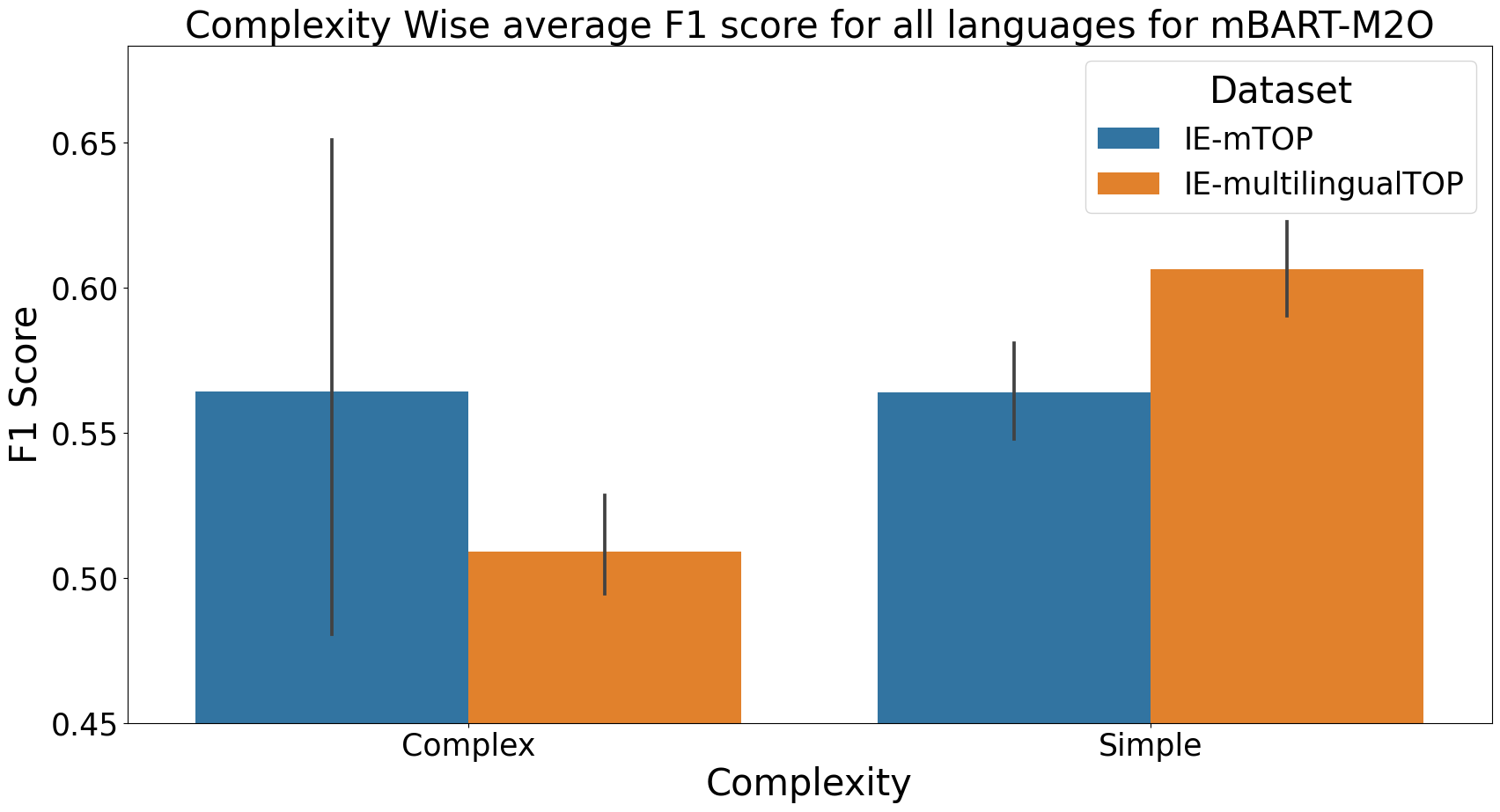}}
    \caption{\small Complexity Wise all language average F1 score in \itop dataset for \itop and \indictop for \mbarttr. 
    }
    \label{fig: compexity f1}
    \vspace{-0.5em}
\end{figure}

\paragraph{Logical Form Complexity:} We evaluate the performance of the \mbarttr model on utterances with simple and complex logical form structures in the Train All setting for \itop and \indictop datasets. Simple utterances have a flat representation with a single intent, while complex utterances have multiple levels \footnote{depth $>=$ 2} of branching in the parse tree with more than one intent. In \indicatis, instances are only attributed to simple utterances since they have a single unique intent. Figure \ref{fig: compexity f1} shows, that \mbarttr performs better for complex utterances in \itop, while there is better performance for simple utterances in \indictop due to its larger training data size and a higher proportion of simple logical forms in training data.

\begin{figure}[!h]
    \vspace{-0.5em}
    {\includegraphics[width=7.8cm]{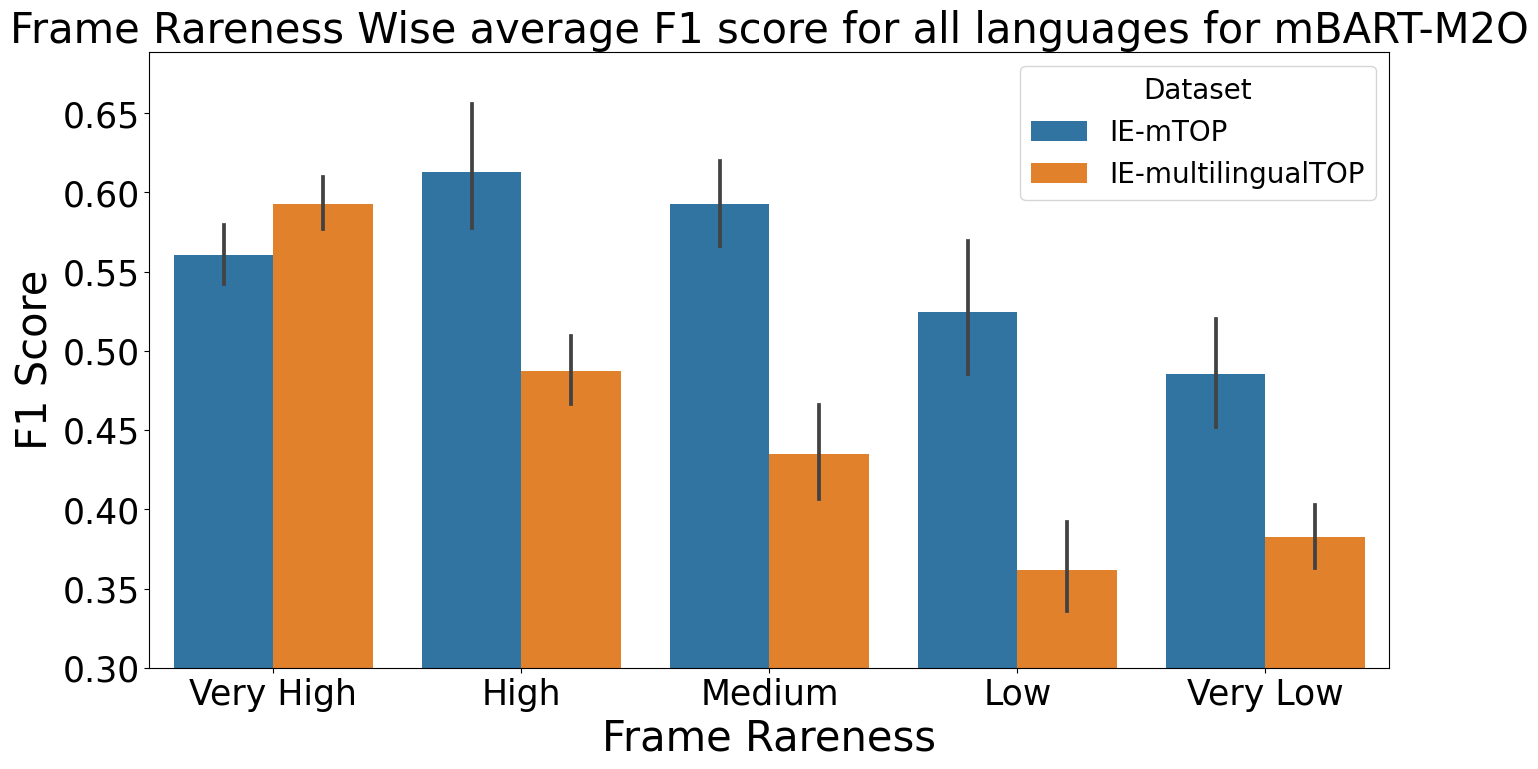}}
    \caption{\small Frame Rareness Wise all language average F1 score in \itop dataset for \itop and \indictop for \mbarttr.  
    }
    \label{fig: frame rareness f1}
    \vspace{-0.50em}
\end{figure}

 \textbf{Effect of Frame Rareness:}
We compared \mbarttr and \indictop on the Train All setting by removing slot values from logical forms and dividing frames into five frequency buckets\footnote{namely very high, high, medium, low and very low.}. A shown in figure \ref{fig: frame rareness f1}, F1 scores increase with frame frequency, and \itop performs better for smaller frequencies while \indictop performs better for very large frequencies. This suggests that \itop has more complex utterances, aiding model learning with limited data, while \indictop's larger training size leads to better performance in very high frequency buckets. 


 \textbf{Post Translation of Slot Values:}
We translate slot values from Hindi to English using IndicTrans for the logical forms of `hi' mTOP and `hi' multi-ATIS++ datasets in the Train All setting. Table \ref{tab: post trans lf} compares the F1 scores of models for \itop and \indicatis datasets, which only had the original Hindi dataset available. Despite minor decreases in scores and visible translation errors, our approach yields accurate translations due to the short length of slot values and the high-resource nature of Hindi. However, we argue that our proposed task or multilingual TOP task is superior in terms of latency and performance, as discussed in \S\ref{sec: task motivation} and \S\ref{sec: result analysis}.

\begin{table}[!h]\centering
\small
\vspace{-0.5em}
\begin{tabular}{llr}\toprule
\textbf{Dataset} &\textbf{Model} &\textbf{F1} \\
\midrule
& {\bf \ibart} &  49 \\
 & {\bf \mbart} &  55 \\
{\bf \itop} & {\bf \mtf} & 50 \\
 & {\bf \ibarttr} & 56 \\
& {\bf \mbarttr} & 58 \\

\hdashline

& {\bf \ibart} & 55 \\
 & {\bf \mbart} & 67 \\
{\bf \indicatis} & {\bf \mtf} & 41 \\
 & {\bf \ibarttr} & 68 \\
& {\bf \mbarttr} & 70 \\
\bottomrule
\end{tabular}
\vspace{-0.5em}
\caption{\small Tree Labelled F1 scores of hindi dataset with post translation of slot values to english for \itop and \indicatis} \label{tab: post trans lf}
\vspace{-0.5 em}
\end{table}

 \textbf{Language Wise Correlation:} We compared the logical form results of each language by calculating the average tree labelled F1 score between the datasets of one language to the other. We then plotted correlation matrices\footnote{for 11 x 11 pairs} and analysed performance on all datasets using \ibarttr and \mbarttr in \textbf{Train All} setting, as described in Figure \ref{fig:consistency_matrix itop}, \ref{fig:consistency_matrix indictop}, and \ref{fig:consistency_matrix indicatis} in Appendix \S\ref{sec: appendix langvslang}.

Our analysis shows that \ibarttr has more consistent predictions than \mbarttr. We also observed that models perform most consistently for the \indicatis dataset. Additionally, related languages, such as `bn' and `as', `mr' and `hi', and `kn' and `te', have high correlation due to script similarity.
\section{Related Work}
\label{sec:relatedework}
\paragraph{Multi-Lingual Semantic Parsing:} Recently, TOP has attracted a lot of attention due to the development of state-of-the-art seq2seq models such as BART \cite{lewis-etal-2020-bart} and T5 \cite{T5}. Moreover, several works have extended TOP to the multilingual setting, such as mTOP, multilingual-TOP, and multi-ATIS++. The recent MASSIVE dataset \cite{fitzgerald2022massive} covers six Indic languages vs eleven in our work, and only contains a flat hierarchical structure of semantic parse. Furthermore, the logical form annotations in MASSIVE are not of a similar format to those in the standard TOP dataset.

\paragraph{IndicNLP:} Some works have experimented with code-mixed Hindi-English utterances for semantic parsing tasks, such as CST5 \cite{hinglishtop}. In addition to these advances, there have been significant contributions to the development of indic-specific resources for natural language generation and understanding, such as IndicNLG Suite \citet{indicnlg}, IndicBART \citet{dabre-etal-2022-indicbart}, and IndicGLUE \citet{kakwani-etal-2020-indicnlpsuite}. Also, some studies have investigated the intra-bilingual setting for multilingual NLP tasks, such as IndicXNLI \cite{aggarwal-indicxnli} and EI-InfoTabs \cite{agarwal-etal-2022-bilingual}. In contrast to prior works, we focus on the complex structured semantic parsing task.

\paragraph{LLMs and Zero Shot:} Our work is also related to zero-shot cross-lingual \citep{sherborne-lapata-2022-zero} and cross-domain \citep{liu-etal-2021-x2parser} semantic parsing, which aims to parse utterances in unseen languages or domains. Moreover, recent methods use scalable techniques such as automatic translation and filling \citep{nicosia-etal-2021-translate-fill} and bootstrapping with LLMs \citep{awasthi2023bootstrapping, amazon-linguist, Scao2022BLOOMA1} to create semantic parsing datasets without human annotation. Unlike previous methods such as Translate-Align-Project (TAP) \cite{tap} and Translate and Fill (TAF) \cite{nicosia-etal-2021-translate-fill}, which generate semantic parses of translated sentences, they propose a novel approach that leverages LLMs to generate semantic parses of multilingual utterances.

\vspace{-.5em}
\section{Conclusion and Future Work}
\label{sec:conclusion}
\vspace{-.5em}

We present a unique inter-bilingual semantic parsing task, and publish the \datasetName suite, which consists of 3 inter-bilingual semantic parsing datasets for 11 Indic languages. Additionally, we discuss the advantages of our proposed approach to semantic parsing over prior methods. We also analyze the impact of various models and train-test procedures on \datasetName performance. Lastly, we examine the effects of variation in logical forms and languages on model performance and the correlation between languages. 




For future work, we plan to release a SOTA model, explore zero-shot parsing \cite{sherborne-lapata-2022-zero}, enhance \datasetName with human translation \cite{nllb}, explore zero-shot dataset generation \cite{nicosia-etal-2021-translate-fill}, leverage LLM for scalable and diverse dataset generation\cite{amazon-linguist, awasthi2023bootstrapping}, and evaluate instruction fine-tuning models.

\section{Limitations}

One of the main limitations of our approach is the use of machine translation to create the \datasetName suite. However, we showed that the overall quality of our dataset is comparable to Samanantar, a human-verified translation dataset. Furthermore, previous studies \citet{google-next-thousand, huang-1990-machine, moon-etal-2020-revisiting, quality-estimation-round-trip} have shown the effectiveness of quality estimation in referenceless settings. Lastly, we have also extensively evaluated our dataset with the help of 3 human evaluators for each language as described in \S\ref{sec: data creation}. We can further take help of GPT4 in future to evaluate the translations in a scaled manner \cite{gilardi2023chatgpt}. 


The second point of discussion focuses on the motivation for preserving logical form slot values in English. We explore the use cases where querying data in English is crucial, and how this approach can enhance models by reducing latency, limiting vocabulary size, and handling system redundancy. While open-source tools currently cannot achieve this, it would be valuable to evaluate the effectiveness of this task by comparing it with the other two discussed approaches. To accomplish this, we suggest using a dialogue manager and scoring the performance of its responses on the three TOP approaches outlined in the paper.


Another potential limitation of our dataset is that it may contain biases and flaws inherited from the original TOP datasets. However, we contend that spoken utterances are generally simpler and more universal than written ones, which mitigates the risk of cultural mismatches in \datasetName dataset. Furthermore, our work is confined only to the Indo-Dravidian Language family of Indic languages due to our familiarity with them and the availability of high-quality resources from previous research. Nonetheless, our approach is easily extendable to other languages with effective translation models, enabling broader applications in various languages worldwide. In the future, we plan to improve our datasets by publicly releasing them through initiatives like NLLB or IndicTransV2, and by collaborating with larger organizations to have the test sets human-translated.

\section{Acknowledgements}
We express our gratitude to Nitish Gupta from Google Research India for his invaluable and insightful suggestions aimed at enhancing the quality of our paper. Additionally, we extend our appreciation to the diligent human evaluators who diligently assessed our dataset. Divyanshu Aggarwal acknowledges all the support from Amex, AI Labs.  We also thank members of the Utah NLP group for their valuable insights and suggestions at various stages of the project; and reviewers their helpful comments. Vivek Gupta acknowledges support from Bloomberg’s Data Science Ph.D. Fellowship.

\bibliography{anthology,custom}
\bibliographystyle{acl_natbib}
\appendix
\appendix
\section{Further Discussions}
\label{sec: appendix further discussions}
\paragraph{\textbf{Why Indic Languages?:}}
Indic languages are a set of Indo-Aryan languages spoken mainly in the Indian subcontinent. These languages combined are spoken by almost 22$\%$ of the total world population in monolingual, bilingual, or multilingual ways. these speakers also are the 2nd largest population of smartphone users, and almost everyone interacts with AI through chatbots. Hence it poses an excellent opportunity for NLP researchers to push state-of-the-art further for standard NLU tasks in these languages to benefit the digital business perspective and make technology more accessible to people through AI. However, most NLU benchmarks lack datasets in those languages despite some being high resource (such as `hi,' `bn,' and `pa'). Moreover, with the introduction of various NLU models like IndicBERT \cite{kakwani-etal-2020-indicnlpsuite}, indicCorp, indicBART \cite{indicnlg}, and state-of-the-art NMT module IndicTrans \cite{ramesh2021samanantar} that has opened new opportunities for researchers to innovate and contribute benchmark datasets which support building NLU models for Indic languages.

Lastly, discourse in languages other than English helps society understand more diverse perspectives and leads to a more inclusive society. As the world is mainly multilingual, various studies have proven that multilingual people can contribute more diverse societal perspectives through digital discourse. 

\paragraph{\textbf{Why IndicTrans translation?}}
Furthermore we use IndicTrans because of the following three reasons, \begin{inparaenum}[(a.)] \item \textbf{Lightweight}: IndicTrans is an extremely lightweight yet state of the art machine translation model for Indic languages. \item \textbf{Indic Coverage}: IndicTrans covers the widest variety of Indic languages as compared to other models like mBART, mT5 and google translate and azure translate are not free for research. \item \textbf{Open Source:} IndicTrans is open source and free for research purposes, more on this is elaborated in \citet{aggarwal-indicxnli}. \end{inparaenum}

\paragraph{\textbf{Why Inter-Bilingual TOP task?}}
Task-Oriented Parsing has seen significant advances in recent years with the rise of attention models in deep learning. There have been significant extensions of this dataset in the form of mTOP \cite{li-etal-2021-mtop} and multilingual-TOP \cite{xia-monti-2021-multilingual}. However, they remain limited in terms of language coverage, only covering a few major global languages and only Hindi in the Indic category.

These datasets are especially difficult to expand to other languages due to the fact that each language has a unique word order and the logical form of each sentence should be modified accordingly. They cannot be altered using a simple dictionary lookup or alignment technique to generate a high-quality dataset. In keeping with this, we propose an inter bilingual TOP task in which only input utterances are translated. As current computers continue to employ English to make decisions and interact with the outside world, modern dialogue managers can work with the logical forms of the English counterparts, construct a response, and translate it back to the input utterance's language.

This resolves the latency issue where the model must first convert the statement to English before parsing it with another seq2seq model. This was mentioned in section \S\ref{sec: result analysis} which demonstrates that end to end models perform better than translate + parsing models in certain instances. Despite the difficulties of learning translation and parsing in a single set of hyper parameters, our research demonstrates that this is feasible with existing seq2seq models, especially models that have being pre-trained with translation task.

\paragraph{\textbf{Task Oriented Parsing in the era of ChatGPT:}}
\label{para: chatgpt}
With the rising popularity of chatGPT \footnote{\url{https://openai.com/blog/chatgpt}} in open-domain conversational AI. It is still a challenge to actually use these large language models in a task-oriented manner. Moreover, these open domain models may not understand the intent of the user correctly or they may take incorrect actions provided a user utterance. These LLMs also have the risk of being biased and toxic. Recent works like HuggingGPT \cite{shen2023hugginggpt} have also shown that while these models may have outstanding language understanding capabilities, it is still better to use task specific models to execute tasks in a narrow scope.

\paragraph{\textbf{Model Coverages:}}
Listed below is the language coverage for all employed multilingual models.

\begin{enumerate} \setlength{\itemsep}{0.5pt}
    \item \textbf{\mbart:} `bn', `gu', `hi',  `ml', `mr', `ta', `te'
    \item \textbf{\mtf:} `bn', `gu', `hi', `kn', `ml', `mr', `pa', `ta', `te'
    \item \textbf{\ibart}: `'as`, bn', `gu', `hi', `kn', `ml', `mr', `or', `pa', `ta', `te'
    \item \textbf{\ibarttr}: `'as`, bn', `gu', `hi', `kn', `ml', `mr', `or', `pa', `ta', `te'
    \item \textbf{\mbarttr}: `bn', `gu', `hi',  `ml', `mr', `ta', `te'
\end{enumerate}

\textbf{Two-step vs End2End parsing:}
\label{sec: end 2 end}
\noindent We measure the translation time of IndicTrans \citep{ramesh2021samanantar} on an NVIDIA T4 GPU and find that it takes 0.015 seconds on average to translate a single utterance from one language to another. In scenario A, this adds 0.03 seconds of latency per utterance, while our approach only adds 0.015 seconds ($\approx \frac{1}{2}$). In scenario B, where the logical form has slot values in Indic, there is no latency overhead for either approach, but there are significant development challenges due to multilingualism as discussed below.

\section{Details: Human Evaluation}
\label{sec: appendix human eval}
\begin{table*}[!htp]\centering
\scriptsize
\begin{tabular}{llrrrrrrrrrrrr}\toprule
\textbf{Dataset} &\textbf{Score} &\textbf{as} &\textbf{bn} &\textbf{gu} &\textbf{hi} &\textbf{kn} &\textbf{ml} &\textbf{mr} &\textbf{or} &\textbf{pa} &\textbf{ta} &\textbf{te} \\
\midrule
& Score$_1$ &3.1 &3 &3.8 &4.3 &3.9 &4.2 &4.1 &4.9 &4.6 &3.8 &4.4 \\
 & Score$_2$ &3 &3 &3.1 &3.7 &3.8 &3.7 &3.5 &4 &4.5 &4.5 &3.5 \\
 & Score$_3$ &3.4 &3.3 &4.1 &4.4 &3.4 &4.5 &4.5 &4.4 &4.3 &3.9 &3.6 \\
 & Pearson$_{1,2}$ &0.8 &0.8 &0.9 &0.8 &0.8 &0.7 &0.6 &0.8 &0.6 &0.7 &0.1 \\
\textbf{\indicatis} & Pearson$_{1,3}$ &0.6 &0.9 &0.2 &0.5 &0.8 &0.7 &0.4 &0.6 &0.7 &0.7 &0 \\
 & Pearson$_{2,3}$ &0.6 &0.8 &0.1 &0.5 &0.6 &0.5 &0.6 &0.7 &0.6 &0.8 &0.7 \\
 & Spearman$_{1,2}$ &0.8 &0.8 &0.8 &0.7 &0.4 &0.5 &0.6 &0.6 &0.3 &0.7 &0.1 \\
 & Spearman$_{1,3}$ &0.7 &0.9 &0.2 &0.5 &0.8 &0.8 &0.5 &0.6 &0.5 &0.7 &0.1 \\
 & Spearman$_{2,3}$ &0.6 &0.9 &0.2 &0.6 &0.3 &0.3 &0.7 &0.6 &0.1 &0.6 &0.7 \\
\hdashline
 & Score$_1$  &2.9 &3 &4 &4.6 &4.4 &4.4 &4.3 &4.9 &4.7 &4.1 &4.4 \\
 & Score$_2$ &3.1 &3.2 &3.7 &4.2 &4.3 &4.2 &4.2 &4.7 &4.5 &4.1 &3.6 \\
 & Score$_3$ &3.2 &3.5 &4 &4.6 &4.3 &3.8 &4.3 &4.7 &4.3 &4.5 &3.5 \\
 & Pearson$_{1,2}$ &0.7 &0.8 &0.5 &0.7 &0.5 &0.7 &0.6 &0.6 &0.7 &0.6 &0.4 \\
\textbf{\indictop} & Pearson$_{1,3}$ &0.6 &0.7 &0.4 &0.5 &0.3 &0.4 &0.7 &0.4 &0.7 &0.4 &0.5 \\
 & Pearson$_{2,3}$ &0.4 &0.8 &0.7 &0.4 &0.6 &0.4 &0.6 &0.2 &0.6 &0.8 &0.9 \\
 & Spearman$_{1,2}$ &0.7 &0.8 &0.4 &0.5 &0.4 &0.5 &0.6 &0.5 &0.5 &0.6 &0.4 \\
 & Spearman$_{1,3}$ &0.6 &0.7 &0.4 &0.3 &0.3 &0.4 &0.7 &0.3 &0.5 &0.3 &0.4 \\
 & Spearman$_{2,3}$ &0.4 &0.8 &0.8 &0.3 &0.6 &0.4 &0.6 &0.1 &0.5 &0.6 &0.7 \\
\hdashline
 & Score$_1$ &2.9 &3.2 &4.2 &4.3 &4.5 &4.3 &4.1 &4.8 &4.7 &4.2 &4.5 \\
 & Score$_2$ &2.8 &3.5 &3.8 &4.2 &4 &3.9 &3.9 &4.4 &4.2 &4 &4.3 \\
 & Score$_3$ &3.2 &3.6 &4 &4.7 &4.3 &3.8 &4 &4.6 &4.4 &4.3 &3.6 \\
 & Pearson$_{1,2}$ &0.8 &0.7 &0.6 &0.7 &0.5 &0.6 &0.8 &0.4 &0.4 &0.4 &0.3 \\
\textbf{\itop} & Pearson$_{1,3}$ &0.6 &0.8 &0.5 &0.4 &0.8 &0.6 &0.7 &0.3 &0.2 &0.4 &0.3 \\
 & Pearson$_{2,3}$ &0.5 &0.7 &0.7 &0.5 &0.5 &0.7 &0.7 &0.6 &0.1 &0.7 &0.6 \\
 & Spearman$_{1,2}$ &0.9 &0.7 &0.6 &0.6 &0.4 &0.6 &0.8 &0.4 &0.3 &0.3 &0.3 \\
 & Spearman$_{1,3}$ &0.6 &0.7 &0.5 &0.3 &0.5 &0.7 &0.6 &0.4 &0.2 &0.3 &0.5 \\
 & Spearman$_{2,3}$ &0.5 &0.7 &0.7 &0.5 &0.3 &0.6 &0.6 &0.7 &0.3 &0.5 &0.4 \\
\bottomrule
\end{tabular}
\caption{\small Detailed Human Evaluation Scores. Score$_x$ refers to the average score of the column language given by x annotator. Pearson$_{x,y}$ refers to the person correlation between the scores of annotators x and y for the column language and similarly for Spearman$_{x,y}$}
\label{tab: appendix human eval detailed}

\end{table*}

In table \ref{tab: appendix human eval detailed} we show the detailed scores of human evaluation process discussed in the main paper \S\ref{sec: data creation}.

\section{Details: Multilingual Models}
\label{sec: appendix model discussion}
\begin{enumerate}
     \item \textbf{Generic Multilingual (Multilingual):} these models are generic Seq2Seq multilingual models, we used \mbart, \mtf \citep{liu-etal-2020-multilingual-denoising, xue-etal-2021-mt5} for experiments for this category.
     \item \textbf{Indic Specific (Indic):} These seq2seq models are specifically pretrained on Indic data, we uexplore \ibart for experiments \citep{dabre-etal-2022-indicbart} in this category.
     \item \textbf{Translation Finetuned (Translation):} These pretrained seq2seq models are finetuned on the translation task with a single target language i.e. English. The models we explored form this category are\ibarttr and \mbarttr \citep{dabre-etal-2022-indicbart, tang-etal-2021-multilingual}.
     \item \textbf{Monolingual (Monolingual):} These seq2seq models are pretrained on English data only. They were  utilize only in the Translate Test setting. The models we explored form this category are \tfl, \tfb \citep{T5} and \bartbase, \bartlarge \citep{lewis-etal-2020-bart}.
 \end{enumerate}
 
\section{Hyperparameters Details}
\label{sec: appendix hyper parameters}

\begin{table*}[!htp]\centering

\scriptsize
\begin{tabular}{lcccccccll}\toprule
\textbf{Hyper Parameter} &\textbf{MS} &\textbf{LR} &\textbf{WD} &\textbf{MSL} &\textbf{BS} & \textbf{NE} &\textbf{PO} &\textbf{PD} \\\midrule
\textbf{\bartbase} &139 &3.00e-3 &0.001 &64 &128 & 50 &Deniosing Autoencoder &Wikepedia Data \cite{lewis-etal-2020-bart} \\
\textbf{\bartlarge} &406 &3.00e-5 &0.001 &64 &16& 50 &Deniosing Autoencoder &Wikepedia Data \\
\textbf{\tfb} &222 &3.00e-3 &0.001 &64 &256& 50 &Multi task Pretraining &C4 \cite{T5}\\
\textbf{\tfl} &737 &3.00e-5 &0.001 &64 &16& 50 &Multi task Pretraining &C4 \\
\textbf{\ibart} &244 &3.00e-3 &0.001 &64 &128& 50 &Deniosing Autoencoder &Indic Corp \cite{kakwani-etal-2020-indicnlpsuite} \\
\textbf{\mbart} &610 &1.00e-4 &0.001 &64 &16& 50 &Deniosing Autoencoder &CC25\cite{liu-etal-2020-multilingual-denoising} \\
\textbf{\mtf} &582 &3.00e-4 &0.001 &64 &16& 50 &Multi task Pretraining &mC4 \cite{xue-etal-2021-mt5}\\
\textbf{\ibarttr} &244 &3.00e-3 &0.001 &64 &128& 50 &Deniosing Autoencoder &PM India \citep{haddow2020pmindia} \\
\textbf{\mbarttr} &610 &1.00e-4 &0.001 &64 &16& 50 &Deniosing Autoencoder &WMT16 \cite{wmt-2020-machine}\\
\bottomrule
\end{tabular}
\caption{\small Hyper Parameters and Pretraining Details}\label{tab: hyper-params}
\end{table*}

In Table \ref{tab: hyper-params} the hyperparamaters are abbreviated as mentioned below:
\begin{enumerate}
    \item \textbf{PO:} Pre-training Objective.
    \item \textbf{PD:} Pretraining Dataset,
    \item \textbf{LR:} Learning Rate,
    \item \textbf{BS:} Batch Size,
    \item \textbf{NE:} Maximum Number of Epochs,
    \item \textbf{WD:} Weight Decay,
    \item \textbf{MSL:} Maximum Sequence Length,
    \item \textbf{MS:} Model Size described as a number of parameters in millions,
    \item \textbf{WS}: Warm-up Step.
\end{enumerate}

All the experiments were run on RTX A5000 GPUs in Jarvis labs \footnote{\url{https://jarvislabs.ai/}}. The code was written in PyTorch and Huggingface accelerate library \footnote{\url{https://huggingface.co/docs/accelerate/index}}. We used early stopping callback in training process with patience of 2 epochs for each setting.

The Average runtime for each for \tfb, \bartbase, \ibart, \ibarttr was 3 minutes for \itop, 1 minute for \indicatis and 5 minutes for \indictop. The Average runtime for each for \tfl, \bartlarge, \mtf ,\mbart, \mbarttr was 5 minutes for \itop, 3 minute for \indicatis and 10 minutes for \indictop.

\section{Vocabulary Augmentation}
\label{sec: appendix vocab augment}
Unique Intents and slots from each dataset (\itop, \indictop, \indicatis) were extracted and added to the tokenizer and model vocabulary so that the models could predict them more accurately. In a typical slot and intent tagging task, these tags would have been treated as classes in the classification model. However, since our models are trained to not predict the entire word but only subwords \cite{T5, lewis-etal-2020-bart} as usually done in modern self-attention architecture \cite{vaswani2017attention}, we decided to include them in the vocabulary so that they can be generated easily during prediction runtime. This also contributed to the reduction of the maximum sequence length to 64 tokens, which improved generalisation as seq2seq models generalise better on shorter sequences \cite{voita-etal-2021-analyzing}. The Excel spreadsheet containing unique slots and intents will be made accessible alongside the code and supplemental materials.

\section{Additional Results}
\subsection{Other Train Test Settings}

We include the results of all other settings except Train All (Already discussed in main paper) in table \ref{tab:itop scores} till \ref{tab: exact English Indic Train}. We have discussed the comparisons of these settings in main paper \S\ref{sec: result analysis}. 
\label{sec: appendix indictop and indicatis}
\begin{table*}[htbp!]
\vspace{-2em}
\centering
\footnotesize
\begin{tabular}
{m{9em}m{10em}<{\raggedright}m{0.5em}m{0.5em}m{0.5em}m{0.5em}m{0.5em}m{0.5em}m{0.5em}m{0.5em}m{0.5em}m{0.5em}m{0.5em}m{5em}<{\centering}}
\toprule \\[-1.2em]
\multirow{2}{*}{\textbf{Dataset}} & \multirow{2}{*}{\textbf{Model}} & \multicolumn{11}{c}{\textbf{Translate Test}} & \multirow{2}{*}{\textbf{ModAvg}}
\\ \\[-1.2em]
&&\textbf{as} &\textbf{bn} &\textbf{gu} &\textbf{hi} &\textbf{kn} &\textbf{ml} &\textbf{mr} &\textbf{or} &\textbf{pa} &\textbf{ta} &\textbf{te} \\
\midrule \\ \\[-2.3em]
 
&\textbf{\bartbase}&28 &37 &35 &\textbf{42} &35 &38 &39 &35 &36 &41 &33 & 36\\
&\textbf{\bartlarge}&30 &41 &38 &44 &38 &41 &41 &39 &38 &\textbf{46} &36 & 39\\
&\textbf{\tfb} &31 &44 &41 &\textbf{49} &41 &43 &43 &41 &42 &47 &41 & 42\\
&\textbf{\tfl}&29 &43 &39 &\textbf{47} &39 &42 &42 &40 &40 &44 &38 & 40\\

\textbf{\itop}&\textbf{\ibart} &30 &40 &36 &42 &36 &40 &39 &38 &37 &\textbf{42} &33 &38 \\

&\textbf{\mtf}&34 &43 &40 &48 &40 &43 &43 &38 &40 &\textbf{45} &38 & 41\\
&\textbf{\mbart}&18 &20 &20 &\textbf{23} &20 &19 &23 &16 &21 &\textbf{23} &21 & 20\\

&\textbf{\ibarttr} &35 &44 &43 &51 &44 &46 &44 &41 &42 &\textbf{49} &41 & 44\\
&\textbf{\mbarttr} &36 &45 &45 &\textbf{50} &45 &47 &46 &41 &46 &53 &43 & \textbf{45}\\


\hdashline
&\textbf{Language Average} &30 &40 &37 &\textbf{44} &38 &40 &40 &37 &38 &43 &36 &38 \\

\midrule

&\textbf{\bartbase}&11 &15 &\textbf{16} &\textbf{16} &13 &14 &13 &14 &14 &14 &16&14 \\
&\textbf{\bartlarge}&12 &18 &19 &\textbf{20} &16 &16 &15 &16 &16 &16 &19 & 17 \\
&\textbf{\tfb}&8 &11 &12 &\textbf{13} &11 &11 &11 &11 &11 &11 &\textbf{13} &11 \\
&\textbf{\tfl}&7 &9 &10 &\textbf{11} &8 &8 &8 &9 &9 &8 &10 & 9\\

\textbf{\indictop}&\textbf{\ibart} &20 &29 &31 &\textbf{32} &27 &29 &25 &26 &27 &25 &31 &27 \\

&\textbf{\mtf}&20 &26 &26 &\textbf{28} &25 &25 &24 &23 &25 &24 &27 &25\\
&\textbf{\mbart}&26 &34 &35 &\textbf{38} &34 &35 &33 &30 &34 &32 &36 &33\\

&\textbf{\ibarttr} &20 &27 &29 &\textbf{30} &27 &28 &25 &25 &26 &25 &29 &26\\
&\textbf{\mbarttr} &30 &42 &45 &\textbf{46} &41 &44 &41 &38 &41 &39 &45 & \textbf{41}\\


\hdashline
& \textbf{Language Average} &17	&23&25&	26&	22&	23&	22&	21&	23&	22&	25&	23 \\

\midrule

&\textbf{\bartbase}& 15 &\textbf{20} &14 &18 &17 &18 &14 &18 &17 &16 &18 &17 \\
&\textbf{\bartlarge}&15 &20 &14 &15 &19 &19 &14 &\textbf{21} &16 &17 &20 &17 \\
&\textbf{\tfb}  &46 &\textbf{70} &52 &62 &61 &65 &47 &51 &58 &51 &66 &57 \\
&\textbf{\tfl}&49 &\textbf{74} &58 &66 &62 &70 &48 &52 &63 &53 &70 &60\\

\textbf{\indicatis}&\textbf{\ibart} &44 &\textbf{66} &46 &56 &54 &63 &47 &46 &58 &49 &63 &54\\

&\textbf{\mtf} &25 &25 &18 &26 &24 &26 &19 &\textbf{27} &25 &20 &24 &24\\
&\textbf{\mbart}  &55 &70 &58 &70 &66 &\textbf{71} &60 &56 &68 &59 &68 &64\\

&\textbf{\ibarttr} &44 &61 &48 &55 &52 &\textbf{68} &48 &53 &56 &47 &59 &54\\
&\textbf{\mbarttr}&53 &70 &68 &\textbf{76} &67 &73 &63 &62 &69 &56 &71 &\textbf{66} \\

\hdashline
& \textbf{Language Average} &38 &\textbf{53} &42 &49 &47 &\textbf{53} &40 &43 &48 &41 &51 &46 \\
\bottomrule
\end{tabular}

\vspace{-0.5em}
\caption{\footnotesize 
   $Tree$\_$Labelled$\_$F1*100$ scores for the all the dataset for \textbf{Translate Test} settings. \textbf{ModAvg} is shorthand for Model Average. The bold numbers in the table indicate the row-wise maximum, i.e. the model's best language performance in the given context. The numbers in bold in the \textbf{ModAvg} column indicate the model with the best performance for the train-test strategy specified in the table's heading. Similarly, the numbers in bold in the \textbf{Language Average} row indicate the language with the best performance for that train-test strategy. 
}
\label{tab:itop scores}
\end{table*}
\begin{table*}[htbp!]
\centering
\footnotesize
\vspace{-3em}
\begin{tabular}{m{9em}<{\raggedright}m{10em}<{\raggedright}m{0.5em}m{0.5em}m{0.5em}m{0.5em}m{0.5em}m{0.5em}m{0.5em}m{0.5em}m{0.5em}m{0.5em}m{0.5em}m{7.5em}<{\centering}}
\toprule \\[-1.2em]
\multirow{2}{*}{\textbf{Dataset}} & \multirow{2}{*}{\textbf{Model}} & \multicolumn{11}{c}{\textbf{Indic Train}} & \multirow{2}{*}{\textbf{Model Average}}
\\ \\[-1.2em]

&&\textbf{as} &\textbf{bn} &\textbf{gu} &\textbf{hi} &\textbf{kn} &\textbf{ml} &\textbf{mr} &\textbf{or} &\textbf{pa} &\textbf{ta} &\textbf{te} \\
\midrule \\ \\[-2.3em]
&\textbf{\ibart} &19 &\textbf{55} &35 &53 &33 &30 &50 &15 &31 &45 &44 &37 \\
&\textbf{\mbart} &41 &51 &14 &\textbf{60} &22 &25 &25 &4 &44 &0 &57 &31  \\
\textbf{\itop} &\textbf{\mtf} &30 &22 &28 &52 &50 &\textbf{54} &36 &8 &36 &53 &15 &35  \\
&\textbf{\ibarttr} &50 &55 &45 &61 &55 &58 &58 &53 &13 &56 &\textbf{59} &51 \\
&\textbf{\mbarttr} &55 &59 &61 &\textbf{66} &56 &63 &57 &52 &53 &59 &63 &\textbf{59} \\
\hdashline
& \textbf{Language Average} &39 &48 &37 &\textbf{58} &43 &46 &45 &26 &35 &43 &48 &43 \\

\midrule
&\textbf{\ibart} &36 &29 &24 &\textbf{65} &48 &9 &56 &30 &37 &42 &40 &38 \\
&\textbf{\mbart} &51 &55 &35 &55 &55 &54 &54 &50 &34 &55 &\textbf{57} &50 \\
\textbf{\indictop}&\textbf{\mtf} &45 &\textbf{56} &\textbf{56} &20 &23 &49 &47 &47 &10 &37 &\textbf{56} &41 \\
&\textbf{\ibarttr} &50 &56 &60 &\textbf{63} &60 &20 &55 &15 &57 &57 &62 &50 \\
&\textbf{\mbarttr} &52 &60 &62 &\textbf{65} &60 &59 &57 &57 &51 &58 &64 &\textbf{59} \\
\hdashline
& \textbf{Language Average} &47 &51 &47 &\textbf{54} &49 &38 &\textbf{54} &40 &38 &50 &56 &48 \\
\midrule

&\textbf{\ibart} &12 &16 &8 &\textbf{25} &15 &19 &22 &22 &23 &22 &18 &19 \\
&\textbf{\mbart} &16 &18 &10 &30 &10 &10 &18 &13 &\textbf{33} &20 &15 &18 \\
\textbf{\indicatis}&\textbf{\mtf} &15 &\textbf{39} &16 &18 &24 &18 &25 &6 &11 &35 &28 &22 \\
&\textbf{\ibarttr} &34 &\textbf{86} &63 &68 &73 &74 &57 &63 &64 &63 &71 &68 \\
&\textbf{\mbarttr} &71 &\textbf{92} &82 &81 &69 &80 &72 &4 &66 &74 &82 &\textbf{70} \\
\hdashline
& \textbf{Language Average} &30 &\textbf{50} &36 &44 &38 &40 &39 &22 &39 &43 &43 &39 \\
\bottomrule
\end{tabular}

\caption{\footnotesize 
   $Tree$\_$Labelled$\_$F1*100$ scores for the all the dataset for \textbf{Indic Train} setting.  The numbers in bold in the \textbf{Model Average} column indicate the model with the best performance for the train-test strategy specified in the table's heading. Similarly, the numbers in bold in the \textbf{Language Average} row indicate the language with the best performance for that train-test strategy. 
}
\label{tab: indictop scores}
\end{table*}
\begin{table*}[htbp!]
\centering
\footnotesize
\vspace{-3em}

\begin{tabular}{m{9em}<{\raggedright}m{10em}<{\raggedright}m{0.5em}m{0.5em}m{0.5em}m{0.5em}m{0.5em}m{0.5em}m{0.5em}m{0.5em}m{0.5em}m{0.5em}m{0.5em}m{7.5em}<{\centering}}
\toprule \\[-1.2em]
\multirow{2}{*}{\textbf{Dataset}} & \multirow{2}{*}{\textbf{Model}} & \multicolumn{11}{c}{\textbf{English+Indic Train}} & \multirow{2}{*}{\textbf{Model Average}}
\\ \\[-1.2em]
& &\textbf{as} &\textbf{bn} &\textbf{gu} &\textbf{hi} &\textbf{kn} &\textbf{ml} &\textbf{mr} &\textbf{or} &\textbf{pa} &\textbf{ta} &\textbf{te} \\
\midrule \\ \\[-2.3em]

&\textbf{\ibart} &34 &37 &42 &\textbf{58} &41 &35 &54 &10 &42 &44 &43 &40 \\
&\textbf{\mbart} &50 &52 &58 &56 &54 &51 &55 &0 &42 &\textbf{59} &57 &49 \\
\textbf{\itop}&\textbf{\mtf} &31 &25 &45 &\textbf{60} &48 &36 &44 &21 &6 &46 &48 &37 \\
&\textbf{\ibarttr} &51 &54 &57 &60 &57 &58 &54 &57 &57 &55 &\textbf{62} &57 \\
&\textbf{\mbarttr} & 57 &60 &60 &65 &62 &\textbf{66} &58 &55 &58 &65 &64 &\textbf{61}  \\
\hdashline
& \textbf{Language Average} &45 &46 &52 &\textbf{60} &52 &49 &53 &29 &41 &54 &55 &49 \\
\midrule

&\textbf{\ibart} &43 &45 &52 &53 &47 &40 &\textbf{57} &30 &47 &38 &49 &46 \\
&\textbf{\mbart} &0 &35 &35 &39 &0 &56 &48 &22 &58 &0 &\textbf{60} & 32\\
\textbf{\indictop}&\textbf{\mtf} &14 &53 &\textbf{56} &50 &53 &50 &50 &48 &52 &51 &\textbf{56} &48\\
&\textbf{\mbarttr} &56 &60 &63 &\textbf{66} &61 &60 &57 &57 &60 &60 &64 &60\\
&\textbf{\ibarttr} &54 &56 &60 &\textbf{63} &60 &58 &54 &57 &24 &57 &\textbf{63} &\textbf{55}\\
\hdashline
& \textbf{Language Average} &33 &50 &53 &\textbf{54} &44 &53 &53 &43 &48 &41 &58 &48 \\

\midrule

&\textbf{\ibart} &34 &12 &12 &\textbf{58} &25 &21 &65 &12 &30 &16 &37 &29 \\
&\textbf{\mbart} &43 &22 &69 &\textbf{78} &14 &54 &58 &12 &36 &10 &66 &42 \\
\textbf{\indicatis}&\textbf{\mtf} &25 &36 &28 &38 &33 &\textbf{44} &23 &23 &35 &30 &35 &32 \\
&\textbf{\mbarttr} &21 &\textbf{86} &78 &74 &73 &76 &56 &64 &72 &65 &75 &\textbf{67} \\
&\textbf{\ibarttr} &71 &\textbf{87} &77 &77 &71 &82 &74 &54 &45 &71 &82 &72 \\
\hdashline
& \textbf{Language Average} &39 &49 &53 &\textbf{65} &43 &55 &55 &33 &44 &38 &59 &48 \\
\bottomrule
\end{tabular}

\caption{\footnotesize 
   $Tree$\_$Labelled$\_$F1*100$ scores for the all the dataset for \textbf{English+Indic Train} setting.  The numbers in bold in the \textbf{Model Average} column indicate the model with the best performance for the train-test strategy specified in the table's heading. Similarly, the numbers in bold in the \textbf{Language Average} row indicate the language with the best performance for that train-test strategy. 
}
\label{tab: english indic scores}
\end{table*}

\subsection{Translate Test vs End2End models}
While the performance of Monolingual models in the Translate Test setting is adequate, the performance of models in the end-to-end Train All setting outperform. Translation is prone to error, and the acquired logical form in English cannot be guaranteed to be precise. Moreover, a two-step approach to translation followed by parsing will incur greater execution time than a unified model.

\subsection{Unified Models Results}
\label{sec: appendix unified}

In unified models, we observe a gain of atleast 0.15 in all languages for all datasets for both \ibarttr and \mbarttr.  

\subsection{Language verses Language}
\label{sec: appendix langvslang}

\begin{figure*}
    \vspace{-3em}
    \subfloat[\ibarttr]{\includegraphics[scale=0.6]{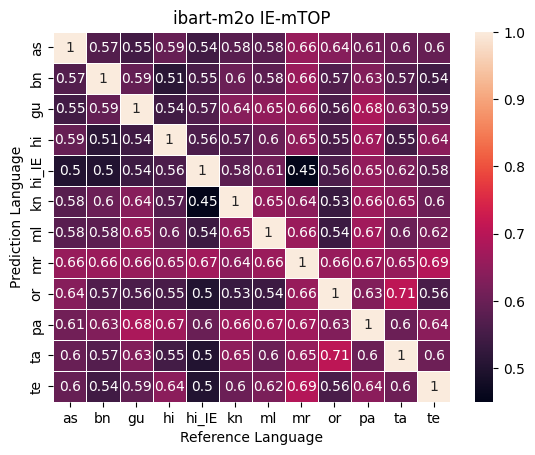}}
    \subfloat[\mbarttr]{\includegraphics[scale=0.6]{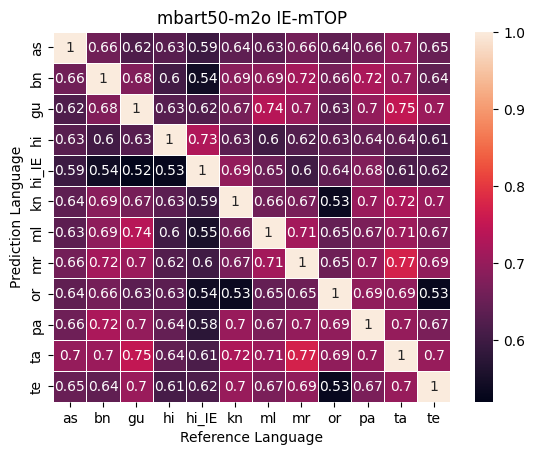}}
   
    \caption{\small Language wise f1 score of predictions of 2 languages for \textbf{\itop dataset} for \textbf{Train All} setting}
    \label{fig:consistency_matrix itop}
    \vspace{-1em}
\end{figure*}

\begin{figure*}
    \vspace{-3em}
    \subfloat[\ibarttr]{\includegraphics[scale=0.6]{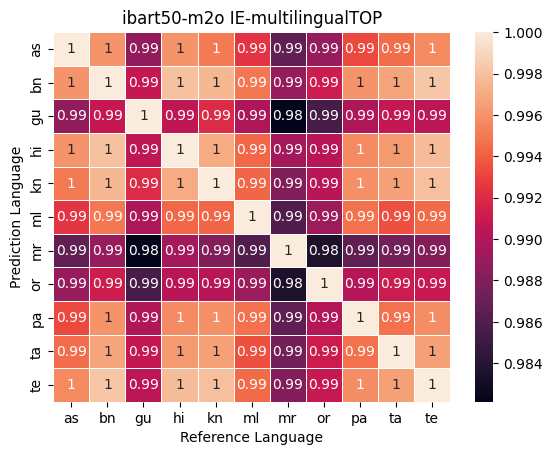}}
    \subfloat[\mbarttr]{\includegraphics[scale=0.6]{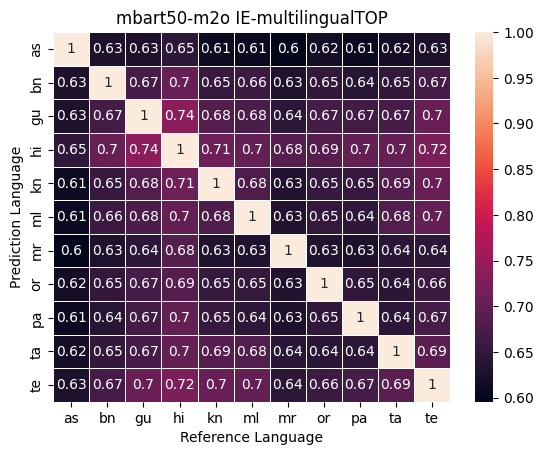}}
   
    \caption{\small Language wise f1 score of predictions of 2 languages for \textbf{\indictop Dataset} for \textbf{Train All} settings}
    \label{fig:consistency_matrix indictop}
\end{figure*}

\begin{figure*}
    \vspace{-3em}
    \subfloat[\ibarttr]{\includegraphics[scale=0.6]{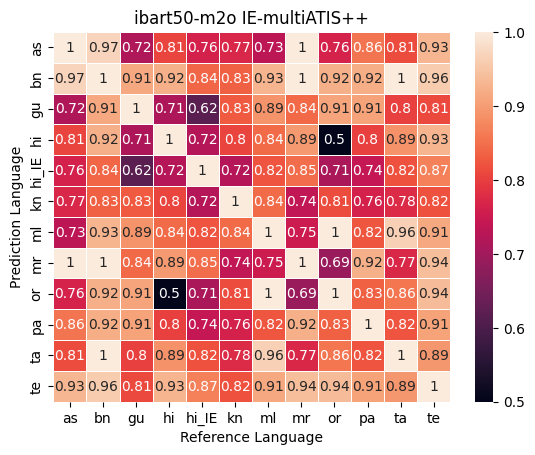}}
    \subfloat[\mbarttr]{\includegraphics[scale=0.6]{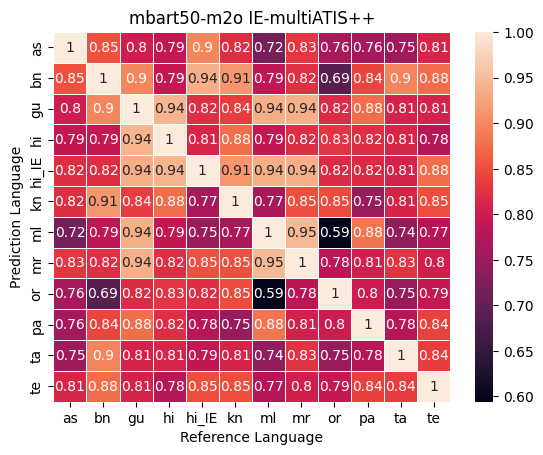}}
   
    \caption{\small Language wise f1 score of predictions of 2 languages for \textbf{\indicatis Dataset} for \textbf{Train All} settings}
    \label{fig:consistency_matrix indicatis}
\end{figure*}

From figure \ref{fig:consistency_matrix itop},  \ref{fig:consistency_matrix indictop}, \ref{fig:consistency_matrix indicatis} we observe that \ibarttr is a more consistent than \mbarttr. 

\subsection{Exact Match Results}
\label{sec: appendic exact match}
\begin{table*}[!htpb]
\vspace{-1.em}
\centering
\footnotesize
\begin{tabular}{m{9em}<{\raggedright}m{10em}<{\raggedright}m{0.4em}m{0.4em}m{0.4em}m{0.4em}m{0.4em}m{0.4em}m{0.4em}m{0.4em}m{0.4em}m{0.4em}m{0.4em}m{0.75em}m{0.75em}m{4.5em}<{\centering}}
\toprule \\[-1.2em]
\multirow{2}{*}{\textbf{Dataset}} & \multirow{2}{*}{\textbf{Model}} & \multicolumn{13}{c}{\textbf{Train All}} & \multirow{2}{*}{\textbf{ModAvg}}
\\ \\[-1.2em]

&&\textbf{as} &\textbf{bn} &\textbf{gu} &\textbf{hi} &\textbf{kn} &\textbf{ml} &\textbf{mr} &\textbf{or} &\textbf{pa} &\textbf{ta} &\textbf{te} &\textbf{hi$_{O}$} &\textbf{hi$_{IE}$}  \\
\midrule \\ \\[-2.3em]

\multirow{6}{*}{\textbf{\itop}} &\textbf{\ibart} &31 &32 &29 &\textbf{42} &29 &32 &42 &20 &28 &30 &31 &64 &49 &35 \\
&\textbf{\ibarttr} &42 &40 &46 &48 &46 &52 &47 &47 &48 &48 &\textbf{50} &68 &53 &49 \\
&\textbf{\mbart} &37 &33 &40 &\textbf{48} &39 &42 &38 &43 &36 &42 &35 &62 &51 &42 \\
&\textbf{\mbarttr} &48 &45 &50 &50 &50 &53 &49 &50 &47 &\textbf{53} &51 &67 &54 &\textbf{51} \\
&\textbf{\mtf} &43 &47 &51 &\textbf{52} &50 &51 &50 &50 &47 &51 &\textbf{52} &59 &55 &\textbf{51} \\
\hdashline
&\textbf{Language Average} &40 &39 &43 &\textbf{46} &43 &\textbf{46} &45 &42 &41 &45 &44 &61 &50 &45 \\
\midrule

\multirow{6}{*}{\textbf{\indictop}} &\textbf{\ibart} &35 &38 &42 &\textbf{56} &39 &37 &47 &22 &38 &36 &43 &\NA &\NA &39 \\
&\textbf{\ibarttr} &45 &47 &47 &55 &46 &46 &52 &45 &53 &50 &\textbf{57} &\NA &\NA &49 \\
&\textbf{\mbart} &37 &41 &43 &\textbf{48} &41 &41 &36 &40 &40 &41 &47 &\NA &\NA &41 \\
&\textbf{\mbarttr} &49 &53 &55 &\textbf{60} &53 &53 &48 &52 &52 &53 &59 &\NA &\NA &\textbf{53} \\
&\textbf{\mtf} &43 &49 &52 &\textbf{56} &52 &50 &47 &45 &49 &48 &54 &\NA &\NA &50 \\
\hdashline
&\textbf{Language Average} &28 &31 &33 &\textbf{37} &32 &31 &31 &27 &30 &30 &34 &\NA &\NA &31 \\
\midrule

\multirow{6}{*}{\textbf{\indicatis}} &\textbf{\ibart} &37 &20 &23 &\textbf{41} &32 &23 &37 &13 &39 &38 &19 &34 &16 &29 \\
&\textbf{\ibarttr} &43 &45 &40 &\textbf{59} &53 &44 &58 &34 &45 &46 &40 &55 &37 &46 \\
&\textbf{\mbart} &60 &85 &73 &\textbf{76} &75 &76 &60 &59 &67 &66 &72 &36 &18 &63 \\
&\textbf{\mbarttr} &67 &80 &71 &\textbf{73} &71 &71 &66 &58 &72 &66 &68 &49 &31 &\textbf{65} \\
&\textbf{\mtf} &45 &70 &58 &\textbf{61} &60 &\textbf{61} &45 &44 &52 &51 &57 &34 &16 &50 \\
\hdashline
&\textbf{Language Average} &50 &60 &53 &\textbf{62} &58 &55 &53 &42 &55 &53 &51 &42 &24 &51 \\

\bottomrule
\end{tabular}
\caption{\footnotesize 
   $Exact$\_$Match$$*100$ scores for the all the dataset for \textbf{Train All} settings. \textbf{ModAvg} is shorthand for Model Average. The bold numbers in the table indicate the row-wise maximum, i.e. the model's best language performance in the given context. The numbers in bold in the \textbf{ModAvg} column indicate the model with the best performance for the train-test strategy specified in the table's heading. Similarly, the numbers in bold in the \textbf{Language Average} row indicate the language with the best performance for that train-test strategy.}
   \label{tab: exact match train all}
\end{table*}

\begin{table*}[!htpb]\centering
\vspace{-1.em}
\footnotesize
\begin{tabular}{m{9em}<{\raggedright}m{10em}<{\raggedright}m{0.4em}m{0.4em}m{0.4em}m{0.4em}m{0.4em}m{0.4em}m{0.4em}m{0.4em}m{0.4em}m{0.4em}m{0.4em}m{8em}<{\centering}}
\toprule \\[-1.2em]
\multirow{2}{*}{\textbf{Dataset}} & \multirow{2}{*}{\textbf{Model}} & \multicolumn{11}{c}{\textbf{Translate Test}} & \multirow{2}{*}{\textbf{Model Average}}
\\ \\[-1.2em]

&&\textbf{as} &\textbf{bn} &\textbf{gu} &\textbf{hi} &\textbf{kn} &\textbf{ml} &\textbf{mr} &\textbf{or} &\textbf{pa} &\textbf{ta} &\textbf{te}  \\
\midrule \\ \\[-2.3em]

\multirow{10}{*}{\textbf{\itop}} &\textbf{\ibart} &29 &40 &38 &\textbf{47} &38 &40 &41 &39 &37 &43 &34 &\textbf{39} \\
&\textbf{\ibarttr} &28 &37 &36 &\textbf{46} &37 &39 &39 &39 &35 &43 &35 &38 \\
&\textbf{\bartbase} &18 &28 &28 &\textbf{35} &27 &29 &29 &29 &28 &33 &24 &28 \\
&\textbf{\bartlarge} &23 &35 &33 &40 &33 &36 &36 &36 &33 &\textbf{41} &30 &34 \\
&\textbf{\mbart} &13 &14 &15 &\textbf{17} &15 &13 &18 &15 &16 &16 &14 &15 \\
&\textbf{\mbarttr} &29 &38 &39 &44 &38 &39 &39 &36 &38 &\textbf{46} &36 &38 \\
&\textbf{\mtf} &26 &36 &33 &\textbf{42} &33 &36 &36 &33 &32 &38 &31 &34 \\
&\textbf{\tfb} &21 &33 &31 &\textbf{40} &30 &31 &33 &35 &31 &37 &32 &32 \\
&\textbf{\tfl} &20 &33 &29 &\textbf{38} &29 &31 &32 &35 &30 &35 &29 &31 \\
\hdashline
&\textbf{Language Average} &23 &33 &31 &\textbf{39} &31 &33 &34 &33 &31 &37 &29 &32 \\
\midrule

\multirow{10}{*}{\textbf{\indictop}} &\textbf{\ibart} &16 &24 &26 &\textbf{28} &21 &24 &20 &21 &22 &20 &26 &23 \\
&\textbf{\ibarttr} &13 &20 &23 &\textbf{24} &20 &21 &18 &19 &19 &19 &22 &20 \\
&\textbf{\bartbase} &12 &13 &13 &\textbf{14} &11 &12 &11 &11 &12 &11 &13 &12 \\
&\textbf{\bartlarge} &10 &15 &16 &\textbf{17} &13 &14 &12 &14 &13 &14 &16 &14 \\
&\textbf{\mbart} &22 &30 &31 &\textbf{35} &30 &31 &29 &26 &29 &28 &32 &29 \\
&\textbf{\mbarttr} &26 &38 &40 &\textbf{43} &36 &38 &36 &33 &35 &34 &40 &\textbf{36} \\
&\textbf{\mtf} &15 &20 &21 &\textbf{23} &19 &20 &18 &18 &20 &19 &21 &19 \\
&\textbf{\tfb} &12 &13 &12 &\textbf{15} &10 &12 &13 &9 &11 &14 &14 &12 \\
&\textbf{\tfl} &22 &23 &22 &25 &26 &26 &25 &26 &26 &26 &\textbf{27} &25 \\
\hdashline
&\textbf{Language Average} &16 &22 &23 &\textbf{25} &21 &22 &20 &20 &21 &21 &23 &21 \\
\midrule

\multirow{10}{*}{\textbf{\indicatis}} &\textbf{\ibart} &30 &49 &34 &41 &41 &\textbf{51} &34 &33 &43 &33 &44 &39 \\
&\textbf{\ibarttr} &32 &51 &39 &44 &40 &\textbf{59} &37 &42 &43 &35 &46 &43 \\
&\textbf{\bartbase} &31 &\textbf{32} &\textbf{32} &30 &31 &30 &30 &30 &30 &30 &30 &31 \\
&\textbf{\bartlarge} &31 &\textbf{32} &\textbf{32} &30 &31 &30 &30 &30 &31 &31 &31 &31 \\
&\textbf{\mbart} &41 &56 &54 &62 &61 &\textbf{66} &54 &50 &60 &47 &56 &55 \\
&\textbf{\mbarttr} &40 &60 &66 &\textbf{69} &62 &66 &57 &58 &60 &47 &59 &\textbf{59} \\
&\textbf{\mtf} &24 &29 &28 &\textbf{35} &28 &24 &26 &27 &22 &25 &24 &27 \\
&\textbf{\tfb} &34 &53 &44 &48 &\textbf{55} &61 &34 &42 &42 &43 &56 &47 \\
&\textbf{\tfl} &38 &\textbf{60} &51 &57 &56 &68 &34 &42 &50 &44 &57 &51 \\
\hdashline
&\textbf{Language Average} &33 &47 &42 &46 &45 &\textbf{51} &37 &39 &42 &37 &45 &42 \\
\bottomrule
\end{tabular}
\caption{\footnotesize 
   $Exact$\_$Match$$*100$ scores for the all the dataset for \textbf{Translate Test} settings. The bold numbers in the table indicate the row-wise maximum, i.e. the model's best language performance in the given context. The numbers in bold in the \textbf{Model Average} column indicate the model with the best performance for the train-test strategy specified in the table's heading. Similarly, the numbers in bold in the \textbf{Language Average} row indicate the language with the best performance for that train-test strategy. }
   \label{tab: exact match translate test}
\end{table*}

\begin{table*}[htbp!]
\vspace{-1.em}
\centering
\footnotesize
\begin{tabular}{m{9em}<{\raggedright}m{10em}<{\raggedright}m{0.4em}m{0.4em}m{0.4em}m{0.4em}m{0.4em}m{0.4em}m{0.4em}m{0.4em}m{0.4em}m{0.4em}m{0.4em}m{8em}<{\centering}}
\toprule \\[-1.2em]
\multirow{2}{*}{\textbf{Dataset}} & \multirow{2}{*}{\textbf{Model}} & \multicolumn{11}{c}{\textbf{Indic Train}} & \multirow{2}{*}{\textbf{Model Average}}
\\ \\[-1.2em]

&&\textbf{as} &\textbf{bn} &\textbf{gu} &\textbf{hi} &\textbf{kn} &\textbf{ml} &\textbf{mr} &\textbf{or} &\textbf{pa} &\textbf{ta} &\textbf{te}  \\
\midrule \\ \\[-2.3em]

\multirow{6}{*}{\textbf{\itop}} &\textbf{\ibart} &24 &26 &29 &33 &28 &24 &\textbf{44} &12 &25 &23 &23 &26 \\
&\textbf{\ibarttr} &43 &48 &49 &56 &48 &\textbf{53} &52 &47 &6 &49 &50 &46 \\
&\textbf{\mbart} &34 &44 &43 &\textbf{55} &40 &44 &45 &27 &36 &0 &50 &38 \\
&\textbf{\mbarttr} &48 &53 &55 &\textbf{62} &50 &58 &53 &48 &46 &54 &57 &\textbf{53} \\
&\textbf{\mtf} &22 &29 &21 &\textbf{45} &42 &46 &29 &24 &28 &25 &24 &30 \\
\hdashline
&\textbf{Language Average} &34 &40 &39 &\textbf{50} &42 &45 &45 &32 &28 &30 &41 &39 \\
\midrule

\multirow{6}{*}{\textbf{\indictop}} &\textbf{\ibart} &30 &24 &20 &\textbf{61} &43 &37 &51 &25 &31 &37 &32 &36 \\
&\textbf{\ibarttr} &45 &54 &56 &\textbf{60} &56 &15 &51 &20 &54 &53 &59 &48 \\
&\textbf{\mbart} &46 &51 &50 &\textbf{57} &51 &50 &49 &46 &31 &50 &54 &49 \\
&\textbf{\mbarttr} &49 &56 &59 &\textbf{62} &56 &55 &53 &53 &46 &54 &60 &\textbf{55} \\
&\textbf{\mtf} &40 &40 &51 &\textbf{61} &51 &43 &43 &43 &40 &47 &53 &47 \\
\hdashline
&\textbf{Language Average} &42 &45 &47 &\textbf{60} &51 &40 &49 &37 &40 &48 &52 &47 \\
\midrule

\multirow{6}{*}{\textbf{\indicatis}} &\textbf{\ibart} &46 &45 &43 &\textbf{54} &32 &34 &46 &23 &20 &30 &32 &37 \\
&\textbf{\ibarttr} &56 &56 &54 &\textbf{74} &44 &55 &68 &47 &40 &50 &52 &54 \\
&\textbf{\mbart} &56 &67 &\textbf{76} &66 &54 &47 &59 &62 &51 &53 &46 &58 \\
&\textbf{\mbarttr} &66 &\textbf{91} &81 &81 &60 &65 &72 &78 &69 &65 &60 &\textbf{72} \\
&\textbf{\mtf} &46 &53 &47 &\textbf{56} &45 &47 &48 &42 &43 &44 &45 &47 \\
\hdashline
&\textbf{Language Average} &54 &62 &60 &\textbf{66} &47 &50 &59 &50 &45 &48 &47 &53 \\

\bottomrule
\end{tabular}
\caption{\footnotesize 
   $Exact$\_$Match$$*100$ scores for the all the dataset for \textbf{Indic Train} settings. The bold numbers in the table indicate the row-wise maximum, i.e. the model's best language performance in the given context. The numbers in bold in the \textbf{Model Average} column indicate the model with the best performance for the train-test strategy specified in the table's heading. Similarly, the numbers in bold in the \textbf{Language Average} row indicate the language with the best performance for that train-test strategy. }
   \label{tab: exact match Indic Train}
\end{table*}

\begin{table*}[htbp!]
\centering
\footnotesize
\vspace{-1.em}
\begin{tabular}{m{9em}<{\raggedright}m{10em}<{\raggedright}m{0.4em}m{0.4em}m{0.4em}m{0.4em}m{0.4em}m{0.4em}m{0.4em}m{0.4em}m{0.4em}m{0.4em}m{0.4em}m{8em}<{\centering}}
\toprule \\[-1.2em]
\multirow{2}{*}{\textbf{Dataset}} & \multirow{2}{*}{\textbf{Model}} & \multicolumn{11}{c}{\textbf{English+Indic Train}} & \multirow{2}{*}{\textbf{Model Average}}
\\ \\[-1.2em]

&&\textbf{as} &\textbf{bn} &\textbf{gu} &\textbf{hi} &\textbf{kn} &\textbf{ml} &\textbf{mr} &\textbf{or} &\textbf{pa} &\textbf{ta} &\textbf{te}  \\
\midrule \\ \\[-2.3em]

\multirow{6}{*}{\textbf{\itop}} &\textbf{\ibart} &27 &29 &36 &\textbf{53} &34 &28 &49 &17 &34 &37 &36 &35 \\
&\textbf{\ibarttr} &45 &46 &50 &\textbf{54} &51 &53 &50 &53 &53 &51 &\textbf{54} &51 \\
&\textbf{\mbart} &43 &46 &50 &50 &47 &45 &50 &0 &37 &\textbf{54} &50 &43 \\
&\textbf{\mbarttr} &51 &55 &53 &\textbf{61} &56 &62 &54 &51 &53 &60 &\textbf{61} &\textbf{56} \\
&\textbf{\mtf} &23 &30 &37 &\textbf{56} &41 &27 &38 &16 &27 &38 &39 &34 \\
\hdashline
&\textbf{Langauge Average} &38 &41 &45 &\textbf{55} &46 &43 &48 &27 &41 &48 &48 &44 \\
\midrule

\multirow{6}{*}{\textbf{\indictop}} &\textbf{\ibart} &37 &30 &47 &52 &42 &35 &\textbf{53} &25 &42 &33 &44 &40 \\
&\textbf{\ibarttr} &48 &52 &56 &59 &56 &54 &50 &53 &16 &53 &\textbf{60} &51 \\
&\textbf{\mbart} &45 &49 &42 &54 &47 &52 &44 &25 &54 &\textbf{56} &\textbf{56} &48 \\
&\textbf{\mbarttr} &51 &56 &\textbf{59} &63 &57 &56 &53 &53 &56 &57 &61 &\textbf{57} \\
&\textbf{\mtf} &39 &48 &51 &46 &49 &45 &42 &43 &47 &47 &\textbf{52} &46 \\
\hdashline
&\textbf{Language Average} &44 &47 &51 &\textbf{55} &50 &48 &48 &40 &43 &49 &\textbf{55} &48 \\
\midrule

\multirow{6}{*}{\textbf{\indicatis}} &\textbf{\ibart} &28 &32 &32 &\textbf{63} &31 &25 &57 &10 &29 &33 &28 &33 \\
&\textbf{\ibarttr} &74 &\textbf{78} &76 &\textbf{78} &72 &80 &40 &54 &64 &53 &68 &\textbf{67} \\
&\textbf{\mbart} &31 &40 &71 &\textbf{83} &71 &69 &57 &21 &23 &40 &58 &51 \\
&\textbf{\mbarttr} &64 &84 &73 &78 &70 &\textbf{88} &71 &46 &66 &70 &76 &71 \\
&\textbf{\mtf} &18 &25 &22 &\textbf{35} &26 &29 &26 &28 &28 &25 &27 &26 \\
\hdashline
&\textbf{Language Average} &43 &52 &55 &\textbf{67} &54 &58 &50 &32 &42 &44 &51 &50 \\
\bottomrule

\end{tabular}
\caption{\footnotesize 
   $Exact$\_$Match$$*100$ scores for the all the dataset for \textbf{English+Indic Train} settings. The bold numbers in the table indicate the row-wise maximum, i.e. the model's best language performance in the given context. The numbers in bold in the \textbf{Model Average} column indicate the model with the best performance for the train-test strategy specified in the table's heading. Similarly, the numbers in bold in the \textbf{Language Average} row indicate the language with the best performance for that train-test strategy. }
   \label{tab: exact English Indic Train}
\end{table*}

We calculated modified exact match scores as inspired by \citet{awasthi2023bootstrapping} which are agnostic of the positions of the slot tokens in the logical form. These scores are presented in tables \ref{tab: exact match train all}, \ref{tab: exact match translate test}, \ref{tab: exact match Indic Train}, \ref{tab: exact English Indic Train}.  We observed that exact match is a stricter metric as compared to tree labelled F1 \citep{gupta-etal-2018-semantic-parsing}. We also observe that exact match scores are consistent with tree labelled F1 scores across languages, datasets and models.

\section{\textbf{Original verses  Interbilingual Hindi}}
\label{sec: appendix original hindi vs translated hindi results}

As demonstrated by figure \ref{fig: top_vs_btop}, we have data accessible in Hindi for all three settings. To produce Hindi bilingual TOP data, we utilize mTOP and multi-ATIS++ to internally combine Hindi and English data tables by unique id (uid). To construct our dataset, we filter the Hindi utterances column and the English logical form columns; we refer to these datasets as \hiorig in table \ref{tab: main scores}. Furthermore, we conduct tests using original Hindi datasets (slot values in Hindi in logical form) and compare their performance to that of other languages. In the table \ref{tab: main scores}, we refer to these datasets as \himtop for the mTOP dataset and multi-ATIS++ dataset both.

 

\emph{Analysis.} We see a decline in F1 score for all models for \hiorig in both \itop and \indicatis. This might be due to data loss when hindi and english data are combined, as not all utterances of english data are included in both datasets. Furthermore, the hindi utterances in the original dataset may be more complex. The results for \himtop and \hiatis enhances because the tokens were copied from the utterance and the model does not have to transform the tokens to English.

\end{document}